\documentclass[]{uiuc_paper}
\pdfoutput=1

\usepackage[toc,page,header]{appendix}
\usepackage{etoc}

\usepackage{cleveref}
\usepackage{subcaption}
\usepackage{booktabs}
\usepackage{tabularx}
\usepackage{array}
\usepackage{enumitem}
\usepackage{algorithm}
\usepackage{algpseudocode}

\usepackage{latexsym}
\usepackage{url}
\usepackage{amssymb}
\usepackage{amsmath}
\usepackage{amsfonts}
\usepackage[utf8]{inputenc}
\usepackage{booktabs}
\usepackage{pifont}
\usepackage{multirow}
\usepackage{makecell}
\usepackage{xspace}
\usepackage{adjustbox}

\setlist{itemsep=2pt, parsep=0pt, topsep=4pt}
\setlist[itemize]{leftmargin=1.4em}
\setlist[enumerate]{leftmargin=1.6em, label=(\arabic*)}

\newlist{rqlist}{description}{1}
\setlist[rqlist]{font=\sffamily\bfseries, leftmargin=2.7em, labelindent=0pt, labelwidth=2.3em,
  labelsep=0.4em, align=left, itemsep=3pt, parsep=0pt, topsep=4pt}

\newtcolorbox{finding}[2]{%
  enhanced,
  colback=cloudgray, colframe=black!80, boxrule=0.4pt, arc=2mm,
  left=3.5mm, right=3.5mm, top=2mm, bottom=2mm,
  before skip=10pt, after skip=10pt,
  before upper={{\sffamily\bfseries Finding #1.\enspace #2}\par\smallskip}
}

\hypersetup{pdftitle={How Much Can Language Models Gain from Test-Time Computation?}}

\newcommand{\method}[0]{\mbox{SELF-POT}\xspace}

\title{How Much Can Language Models Gain\\from Test-Time Computation?}

\author[1]{Bangji Yang}
\author[2]{Jingyuan Li}
\author[1]{Jiajun Fan}
\author[1]{Yi Evie Zhang}
\author[1]{Ruihan Guo}
\author[3]{Hongbo Ma}
\author[1]{Neil He}
\author[1]{Chumeng Liang}
\author[1]{Qinglong Zheng}
\author[1]{Zhanghan Ni}
\author[1,$\dagger$]{Ge Liu}

\affiliation[1]{University of Illinois at Urbana-Champaign}
\affiliation[2]{University of Washington}
\affiliation[3]{Tsinghua University}

\contribution[$\dagger$]{Corresponding author}

\abstract{How much can test-time computation improve a language model, and at what cost?
Test-time scaling is widely proposed as a substitute for larger models, but existing comparisons mostly evaluate one domain at a time and rarely charge selection to the budget.
We introduce \method, a benchmark and evaluation framework that measures the test-time potential of a model across competition mathematics, competitive programming, and agentic workflows.
\method separates candidate coverage from final accuracy on static tasks, tracks correctness transitions under revision, and measures protocol completion alongside task success in agentic environments.
Under a unified budget rule, it compares Direct inference with parallel sampling and self-revision under fixed multiples of the Direct budget, and charges every model call, including selection and critique, in dollars.
This design supports two kinds of comparison: the gain a model obtains from additional inference, and a lower-cost model with additional inference against a stronger model.
Across five low-cost reasoning models on 350 sealed tasks, with Claude Opus~5.5 Direct as the reference, the returns depend on the domain, the selection rule, and failure handling.
When we replay the retained programming candidate pools, public-example selection raises correct submissions from 376 to 453 of 500 scheduled cells while saving 12--49\% of logical API cost across models, and simply retaining an available candidate when judging fails recovers 61 submissions at unchanged cost.
On identical mathematics pools, judging with fallback yields 186 correct submissions versus 182 for voting, while voting saves 12--21\% of logical API cost.
These controlled replays show how selection and failure handling change the gains realized from the same generated candidates, and they quantify the marginal value of a model judge.
}

\date{\today}
\correspondence{\email{geliu@illinois.edu}}

\begin{document}

\etocdepthtag.toc{mainmatter}

\maketitle

\section{Introduction}
\label{sec:intro}

Test-time scaling (TTS) improves a fixed language model by spending more computation at inference rather than on training.
Parallel methods draw several candidate answers and keep one, chosen by majority vote~\citep{wang2023selfconsistency}, by running candidate programs on tests~\citep{chen2023codet}, or by a model that judges the candidates~\citep{chen2025knockout}.
Sequential methods let the model critique an earlier answer and then revise it~\citep{madaan2023selfrefine}.
When the computation is allocated well, a smaller model can match a much larger one on problems it already solves some of the time~\citep{snell2025scaling}, so test-time scaling looks like a way for a low-cost model to replace a stronger and more expensive one.

For a fixed model and budget, a practitioner must therefore choose among a single attempt, parallel sampling with selection, and self-revision, and the right choice is unlikely to be the same across tasks.
The options differ in what they cost: voting and public-example selection use no additional model tokens, whereas a model judgment can cost as much as another answer.
They also differ in what a task permits: in an app workflow, actions can change the environment, so a flawed attempt cannot simply be discarded and drawn again.

\begin{figure}[t]
\centering
\includegraphics[width=0.94\linewidth]{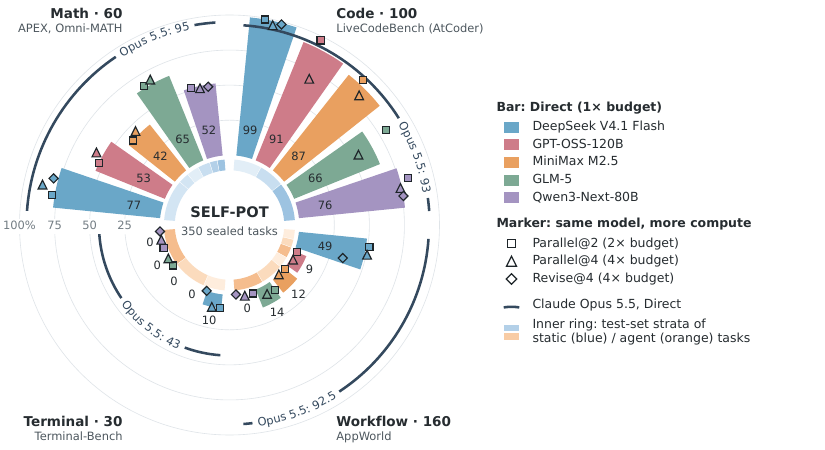}
\caption{\textbf{\method: frozen protocol matrix.} Each quarter (equal angles) is one domain of the 350-task sealed test set, with its sampling strata in the inner ring. Bars show each low-cost model's Direct accuracy (\%), markers the same model under the TTS protocols, and arcs Claude Opus~5.5 answering once. A marker beyond the end of a bar is a gain, and a marker inside a bar is a loss.}
\label{fig:teaser}
\end{figure}

Prior comparisons of test-time scaling methods do not tell a practitioner which option to choose in a given task domain.
Several studies compare parallel and sequential methods under a matched budget~\citep{snell2025scaling,wu2025inference,wang2024tokeneconomies}, but they leave three gaps.
First, most evaluate a single domain, usually mathematics~\citep{snell2025scaling,ghosal2025thinking}, so a method that helps there may or may not help elsewhere.
Second, few charge selection or verification calls to the same budget as generation~\citep{wang2024tokeneconomies,singhi2025solve}, and most count the budget in FLOPs, samples, or tokens rather than in money.
Third, the comparison that motivates test-time scaling, a low-cost model with more computation against a stronger model answering once, has been made only in narrow settings.
\citet{snell2025scaling} and \citet{wu2025inference} compare a smaller and a larger model from the same family (e.g., Llemma-7B and Llemma-34B in the latter) at matched FLOPs and mainly on mathematics.
\citet{stroebl2026limits} compare a weaker and a stronger model on programming tasks.
Neither setting measures the comparison in dollars or across domains.

We therefore ask three questions:
\begin{rqlist}
  \item[RQ1] Under stated budget allowances, which tested parallel and sequential configurations turn extra computation into accuracy, and in which domains?
  \item[RQ2] On static tasks, is opportunity lost in generating a correct candidate, selecting it, or preserving it under revision, and do agent protocols complete their intended execution?
  \item[RQ3] Can a low-cost model with test-time scaling reach the accuracy of a stronger model answering once at no greater cost, and in which cases can it not?
\end{rqlist}

\begin{figure}[t]
\centering
\includegraphics[width=\linewidth,trim=0 0 0 86,clip]{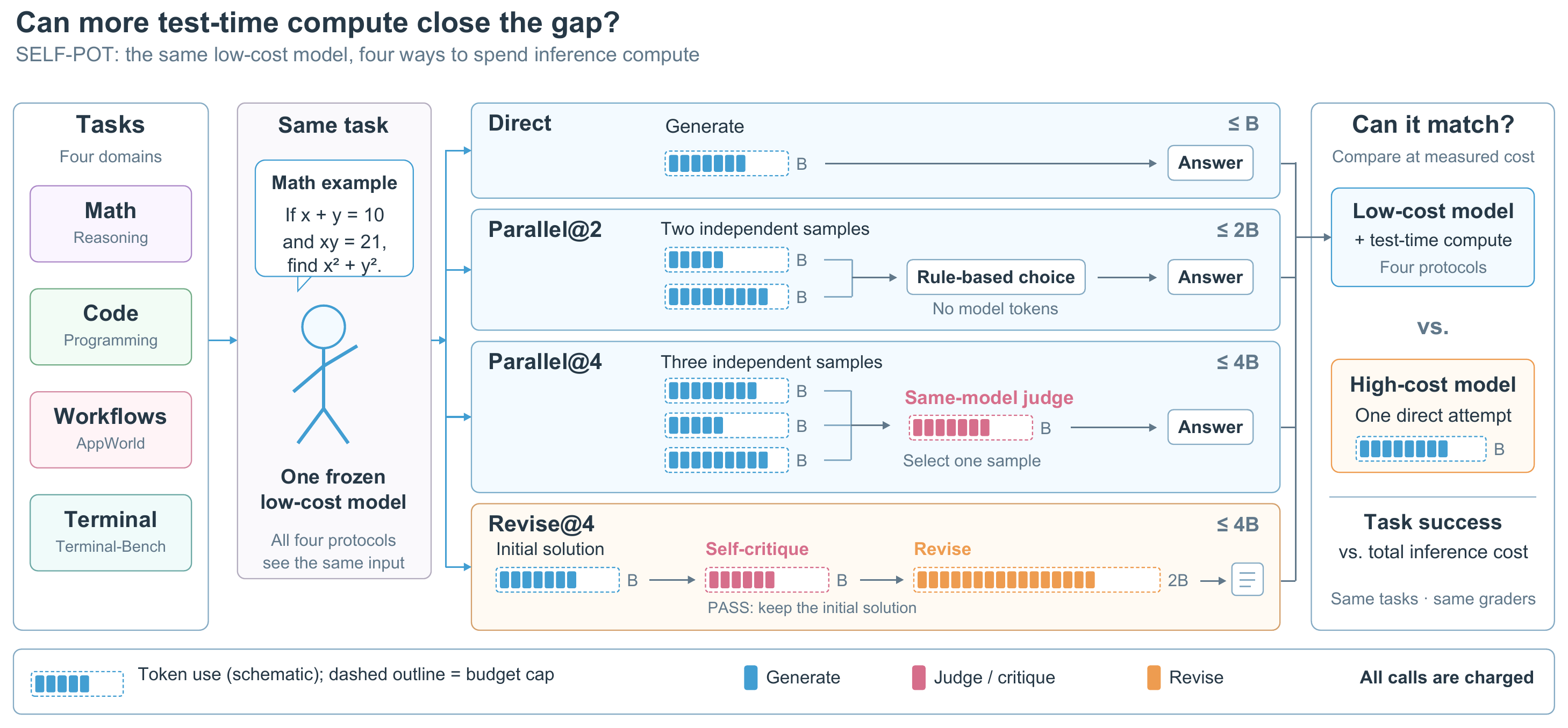}
\caption{\textbf{\method protocols.} One frozen low-cost model answers the same task under four protocols whose total output is capped at a multiple of the budget $B$ of one answer, and every call is charged. On agent tasks, the parallel protocols compare plans that are executed once, and Revise@4 reviews the trajectory every two steps (\S\ref{sec:method-protocols}). The mathematics problem is illustrative.}
\label{fig:overview}
\end{figure}

Answering these questions places four requirements on an evaluation.
It must charge every call, including selection and critique, while keeping the model fixed across roles; it must distinguish the number of candidates, the rule that chooses among them, and the handling of failed stages; its domains must differ in the feedback they provide and in whether actions can be undone; and its test set must be fixed before any test-time call.
\method is built to meet these requirements (Figure~\ref{fig:overview}).
It gives a frozen model a budget of one, two, or four times a single answer and compares answering once (\emph{Direct}) with two parallel protocols and one sequential protocol, each in the simplest form that a reasoning model can run by itself.
\emph{Parallel@2} keeps one of two answers by majority vote in mathematics or by the \emph{public examples} given in a programming task, \emph{Parallel@4} lets the same model read three answers and pick one in a selection stage (\emph{joint selection}), and \emph{Revise@4} critiques the answer and then revises it.
On agent tasks, the parallel protocols compare plans before a single execution (\emph{plan-level parallelism}), and Revise@4 reviews the trajectory every two steps (\emph{periodic self-review}).

We evaluate five low-cost reasoning models on 350 tasks in competition mathematics (APEX and Omni-MATH~\citep{balunovic2025matharena,matharena2025apex,gao2025omnimath}), AtCoder programming from LiveCodeBench~\citep{jain2025livecodebench}, AppWorld workflows~\citep{trivedi2024appworld}, and Terminal-Bench tasks~\citep{tbench2026}, fixed and hashed before any evaluation run (a \emph{sealed} test set), and compare them with Claude Opus 5.5 answering once, the stronger reference model (the \emph{anchor}).
The five models are DeepSeek V4.1 Flash, Qwen3-Next-80B-A3B, GLM-5, GPT-OSS-120B, and MiniMax M2.5.
Beyond the protocol comparison, we replay alternative selection and fallback rules on the saved candidate pools, which separates generating a correct candidate from converting it into a correct submission.
Figure~\ref{fig:teaser} summarizes the frozen protocol matrix, and our main findings are:
\begin{itemize}
  \item \textbf{Failure handling can reverse the measured gain.} Programming joint selection loses 8.6 percentage points (pp) against Direct under the frozen submission rule; retaining the first available candidate when no nonempty choice is returned turns this into a 3.6~pp gain at unchanged cost (\S\ref{sec:results-rq2}).
  \item \textbf{Public evidence can convert more of the same candidates.} On programming, public-example selection yields 453 correct submissions versus 376 for judging, at 12--49\% lower API cost. On identical mathematics pools, judging with fallback reaches 186/292 versus 182/292 for voting (\S\ref{sec:results-rq2}).
  \item \textbf{Agent gains require completed protocols.} Workflow stage failures reach 117/160 for GPT-OSS and 159/160 for MiniMax, so these configurations primarily measure interface reliability; they do not establish the value of completed plan comparison (\S\ref{sec:results-rq2}).
  \item \textbf{Deployment value differs from within-model gain.} Flash Direct remains the coding frontier after the replays are added, and discounted mathematics vote@3 adds a cheaper operating point than judging. Outside programming, Opus remains the most accurate of the tested configurations (\S\ref{sec:results-rq3}).
\end{itemize}

\noindent Our contributions are:
\begin{enumerate}
  \item a measurement protocol that holds the model, information, and grading fixed, caps total output at stated multiples of one attempt, and charges every call (\S\ref{sec:method});
  \item a sealed four-domain test set whose tasks pass a model-independent eligibility check (\S\ref{sec:method-domains});
  \item an evaluation of five low-cost models and a stronger anchor, per model and pooled, with the cost of every configuration (\S\ref{sec:results}); and
  \item controlled candidate-pool replays that separate selection quality, submission failures, and the cost of deployable alternatives (\S\ref{sec:results-rq2}).
\end{enumerate}
We do not propose a new test-time scaling method.
The protocols share one accounting rule, although the feedback available to them and their realized costs differ across domains, and the same-pool replays isolate selection and failure handling on static tasks.

\section{Methodology}
\label{sec:method}

\method measures how much additional inference a model turns into final task success, and whether the resulting accuracy justifies its measured API cost.
The same frozen model supplies every role within a configuration---solver, selector, critic, and agent---and every call it makes is charged.

\subsection{Measured quantity and budget}
\label{sec:method-setup}

We measure $A(\mathcal{M},\pi;\,D,I,E,H,B)$, the expected final success rate of model $\mathcal{M}$ under protocol $\pi$, given the task distribution $D$, the information $I$ visible to the solver, the environment $E$, the harness $H$, and the budget $B$.
We call the gains attainable under these stated conditions the model's \emph{test-time potential}.
The measured rate estimates $A$ for the stated configuration, rather than an intrinsic property of the model.
One execution of a protocol on a task by a model is a \emph{cell}.
For domain $d$, let $B_d$ be the Direct output-token allowance and $m_\pi$ the protocol multiplier.
All model calls $c_1,\ldots,c_T$ in a cell obey
\begin{equation}
  \sum_{t=1}^{T}\tau(c_t)\leq m_\pi B_d,
  \label{eq:budget}
\end{equation}
where $\tau$ counts reasoning and answer tokens.
$B_d$ is 131{,}072 tokens for mathematics and programming and 393{,}216 for an agent episode, and the caps on model calls, tool calls, and wall-clock time scale with $m_\pi$ in the same way (Appendix~\ref{app:budgets}).
These are maximum allowances, not matched realized costs: a protocol may stop early, and provider limits can impose tighter stage caps.
We therefore report accuracy together with realized token use and dollar cost; the configurations do not constitute a full accuracy--budget response curve.

\subsection{Protocols}
\label{sec:method-protocols}

Direct answers once with the allowance $B_d$.
On programming, a Direct attempt may run its program on the public examples before submitting it, and on agent tasks it is a complete multi-call episode, so Direct is neither a short answer nor a single request.
We do not shorten Direct to create headroom for another method.

\paragraph{Static tasks.}
Parallel@2 generates two candidates and keeps one by a rule that uses no model tokens: the majority final answer in mathematics and the most public examples passed in programming.
Ties go to the earlier candidate, so with two parsed answers voting cannot improve on the first; we therefore also compare three-way voting with judging on identical pools (\S\ref{sec:method-decomposition}).
Parallel@4 generates three candidates and allocates another $B$ to a joint-selection stage: the same model reads all three and names one, with public-tool calls available on code.
Revise@4 runs one Self-Refine iteration~\citep{madaan2023selfrefine}.
The model first critiques its answer without writing a new one, a \texttt{PASS} verdict returns the initial answer, and otherwise the model revises it with up to $2B$ tokens.
The suffix denotes a budget cap, not a sample count, so Parallel@4 and Revise@4 share the cap but not their realized cost.
We omit revision at $2B$ because our allocation gives each of its three stages a full allowance (Appendix~\ref{app:protocols}).

\paragraph{Agent tasks.}
\label{sec:method-instantiation}
Actions change the environment, so no agent protocol resets an episode or chooses among executed worlds.
Parallel@$k$ with $k\in\{2,4\}$ drafts $k$ plans from the initial observation, selects one with a same-model call, and executes only that plan in a single episode.
Revise@4 inserts a same-model review after every two executor turns, appends its advice to the history, and continues from the current state.
Planning and plan selection run at low reasoning effort and execution and review at high effort, with all stages sharing the cell's allowance (Appendix~\ref{app:algorithms}).

\subsection{From candidate coverage to final accuracy}
\label{sec:method-decomposition}

For a protocol that selects from a recorded pool of candidates, let $O$ indicate that at least one candidate passes the hidden grader and $S$ that the final submission is correct.
Since the submission must be one of the candidates,
\begin{equation}
  \mathbb{E}[S]=P(O)\,P(S\mid O),
  \label{eq:selection}
\end{equation}
the product of \emph{candidate coverage} and \emph{conversion}.
We partition covered failures into two kinds: the selection returns no valid index (\emph{no choice}), or it indexes a candidate that grades incorrect, including an empty candidate (\emph{wrong choice}).
Coverage uses hidden grades, so it is an offline ceiling for the recorded candidates rather than a deployable oracle.

To isolate conversion, we replay task-appropriate choice rules on the saved Parallel@4 pools: majority voting over three mathematics answers and public-example selection over three programs, each with an earliest-candidate tie rule.
We also assess a deterministic fallback that retains the first nonempty candidate when judging yields no usable answer.
These post-hoc analyses hold generation fixed and are not new equal-cost trials.
Public examples were available to the model judge as well, and hidden grades are used only to score the resulting choices.
Fallback retains the already-incurred judge cost, whereas a rule that omits judging excludes that stage's cost (Appendix~\ref{app:selector-replay}).

For revision, let $p$ be the accuracy of the protocol's own initial answers, $r$ the fraction of initially incorrect submissions that become correct, and $d$ the fraction of correct answers lost in the final submission.
Then
\begin{equation}
  \Delta=(1-p)\,r-p\,d,
  \label{eq:revision}
\end{equation}
so for $r+d>0$ revision gains exactly when $p<r/(r+d)$.
These transitions include empty submissions, and we distinguish output failures from changes to completed answers.
Unchosen agent plans are never executed, so coverage is undefined for agent protocols, which we compare on the same task across separate episodes rather than through repairs within one trajectory.
Both identities are elementary accounting; their value lies in locating where an opportunity is lost.

\subsection{Test set}
\label{sec:method-domains}

The mathematics problems come from APEX~\citep{balunovic2025matharena,matharena2025apex} and Omni-MATH~\citep{gao2025omnimath}, the programming problems from AtCoder through LiveCodeBench v6~\citep{jain2025livecodebench}, and the agent tasks from AppWorld~\citep{trivedi2024appworld} and Terminal-Bench~\citep{tbench2026} (Table~\ref{tab:domains}).
The domains differ in the feedback available while solving and in whether actions can be undone.
Mathematics gives no feedback, programming gives execution on the public examples, and the two agent domains return API responses or shell output from an environment whose changes persist.

\begin{table}[t]
\centering\small
\caption{The four domains of the sealed test set. Hidden answers, tests, and goal checks are used only for grading; Appendix~\ref{app:testset} gives the sampling strata.}
\label{tab:domains}
\begin{tabular*}{\linewidth}{@{\extracolsep{\fill}}lllll@{}}
\toprule
Domain & Source & Tasks & Feedback while solving & Actions reversible \\
\midrule
Math & APEX~\citep{balunovic2025matharena,matharena2025apex}, Omni-MATH~\citep{gao2025omnimath} & 60 & none & yes \\
Code & LiveCodeBench v6~\citep{jain2025livecodebench} (AtCoder) & 100 & public examples and execution & yes \\
Workflow & AppWorld~\citep{trivedi2024appworld} & 160 & app API responses & no \\
Terminal & Terminal-Bench 4.0~\citep{tbench2026} & 30 & shell output & no \\
\bottomrule
\end{tabular*}
\end{table}

Tasks pass a model-independent eligibility check: where a reference solution exists, it must pass and a blank or no-op submission must fail under the pinned runtime.
Every task seen by a model during development was excluded, and the held-out tasks were drawn within difficulty strata with a fixed seed and at most one task per leakage group, then sealed before any test-time call (Appendix~\ref{app:testset}).
No test item was chosen for an observed test-time gain.
Task type, difficulty, feedback, and protocol instantiation change together across domains, so a contrast between domains does not isolate any one of these factors.

\subsection{Grading and cost accounting}
\label{sec:method-accounting}

Mathematics uses a rule-based expression grader, programming runs hidden tests, and workflow and terminal cells use their benchmarks' native final-state evaluators (Appendix~\ref{app:grading}).
An absent or ungradable submission counts as incorrect, including the 15 scheduled cells that have no final record.
For agents, the native grade of the final state is authoritative even when an episode stops at a resource limit, and we report termination status separately.
Selection, critique, planning, review, and execution calls all count toward the cell's allowance and cost.
Identical requests may reuse stored responses across static protocols, but every configuration is charged the logical cost of its own calls.
Reported costs are mean logical API costs over retained cells at the configured rates, not invoices: DeepSeek is charged at its peak rate, with a half-price scenario where it matters, and Bedrock models at list rates (Bedrock prompt-cache discounts apply only to Opus agent requests).
Model outcomes are never retried; bounded infrastructure retries and interface amendments made after sealing are documented in Appendix~\ref{app:accounting}.

\section{Experiments}
\label{sec:results}

\subsection{Setup}
\label{sec:method-models}

\paragraph{Models.}
Five lower-priced reasoning models---DeepSeek V4.1 Flash, Qwen3-Next-80B-A3B, GLM-5, GPT-OSS-120B, and MiniMax M2.5---are compared with Claude Opus~5.5 Direct as the stronger anchor (Table~\ref{tab:models}; Appendix~\ref{app:models}).
``Lower-priced'' refers to token rates, not to cost per solved task.
Only Flash and Qwen3-Next run revision, because the other models' context windows cannot hold a full-budget critique on top of an answer.
GPT-OSS and MiniMax run only Direct on terminal tasks, an adaptive omission after their zero Direct scores; these unrun cells are not counted as failures.
The matrix has 68 configurations and 6{,}180 scheduled cells, of which 6{,}165 have a final record.

\begin{table}[!ht]
\centering\small
\caption{Models. Prices are USD per million input and output tokens on our ledger. Revision eligibility uses the configured input headroom plus a full-budget critique, not a universal context threshold for all revision methods.}
\label{tab:models}
\begin{tabular*}{\linewidth}{@{\extracolsep{\fill}}llrrl@{}}
\toprule
Model & Access & Price (in / out) & Context & Protocols \\
\midrule
DeepSeek V4.1 Flash & DeepSeek API & 0.30 / 1.20\textsuperscript{a} & 1M & all \\
Qwen3-Next-80B-A3B & Amazon Bedrock & 0.15 / 1.20 & 262{,}144 & all \\
GLM-5 & Amazon Bedrock & 1.00 / 3.20 & 202{,}752 & Direct, Parallel \\
GPT-OSS-120B & Amazon Bedrock & 0.1545 / 0.618 & 131{,}072 & Direct, Parallel\textsuperscript{b} \\
MiniMax M2.5 & Amazon Bedrock & 0.30 / 1.20 & 196{,}608 & Direct, Parallel\textsuperscript{b} \\
\midrule
Claude Opus 5.5 (anchor) & Amazon Bedrock & 4.00 / 20.00 & -- & Direct \\
\bottomrule
\end{tabular*}

\smallskip
\raggedright\footnotesize
\textsuperscript{a}Peak rate; the off-peak rate is about half. \textsuperscript{b}Terminal: Direct only, since both models scored 0/30 on terminal Direct and no terminal protocol had improved on Direct for the first three models.
\end{table}

\paragraph{Statistics.}
Each cell is sampled once.
Within a model and domain, we count paired wins (protocol correct, Direct incorrect) and losses and apply an exact two-sided sign test, Holm-corrected across all 44 protocol-versus-Direct comparisons.
For a pooled domain estimate, we average the paired change over the fixed panel of evaluated models and resample tasks within the frozen strata, keeping all models' outcomes on a task together (20{,}000 replicates, pointwise 95\% intervals; Appendix~\ref{app:offline-audit}).

\subsection{RQ1: How gains vary across protocols and domains}
\label{sec:results-rq1}

\begin{table}[t]
\centering
\small
\caption{Frozen protocol matrix: accuracy (Acc) over all scheduled tasks and mean logical API cost in USD over retained cell records. Missing records have no imputed cost. Opus 5.5 is the expensive anchor and runs Direct only. Bold marks the best accuracy among the four displayed protocols in each row; ``--'' means not run; $\ddagger$ includes cells that never finished, counted as incorrect.}
\label{tab:summary}
\setlength{\tabcolsep}{3.7pt}
\begin{tabular}{@{}llrrrrrrrr@{}}
\toprule
 & & \multicolumn{2}{c}{Direct} & \multicolumn{2}{c}{Parallel@2} & \multicolumn{2}{c}{Parallel@4} & \multicolumn{2}{c}{Revise@4} \\
\cmidrule(lr){3-4} \cmidrule(lr){5-6} \cmidrule(lr){7-8} \cmidrule(lr){9-10}
Domain & Model & Acc & \$ & Acc & \$ & Acc & \$ & Acc & \$ \\
\midrule
Math (60) & Opus 5.5 & 0.950 & 0.103 & -- & -- & -- & -- & -- & -- \\
 & DeepSeek V4.1 Flash & 0.767 & 0.051 & 0.783 & 0.098 & \textbf{0.867} & 0.176 & 0.800 & 0.096 \\
 & GPT-OSS-120B & 0.533 & 0.026 & 0.533 & 0.052 & \textbf{0.583}$^\ddagger$ & 0.096 & -- & -- \\
 & MiniMax M2.5 & 0.417 & 0.038 & 0.417 & 0.078 & \textbf{0.450} & 0.151 & -- & -- \\
 & GLM-5 & 0.650 & 0.175 & 0.667$^\ddagger$ & 0.347 & \textbf{0.683}$^\ddagger$ & 0.618 & -- & -- \\
 & Qwen3-Next-80B & \textbf{0.517} & 0.048 & \textbf{0.517} & 0.091 & 0.500 & 0.156 & 0.500 & 0.059 \\
\midrule
Code (100) & Opus 5.5 & 0.930 & 0.028 & -- & -- & -- & -- & -- & -- \\
 & DeepSeek V4.1 Flash & \textbf{0.990} & 0.013 & \textbf{0.990} & 0.027 & 0.960 & 0.077 & 0.980 & 0.017 \\
 & GPT-OSS-120B & 0.910 & 0.025 & \textbf{0.970} & 0.052 & 0.690$^\ddagger$ & 0.081 & -- & -- \\
 & MiniMax M2.5 & 0.870 & 0.039 & \textbf{0.910} & 0.075 & 0.810 & 0.169 & -- & -- \\
 & GLM-5 & 0.660 & 0.103 & \textbf{0.810} & 0.212 & 0.550 & 0.377 & -- & -- \\
 & Qwen3-Next-80B & 0.760 & 0.019 & \textbf{0.820} & 0.038 & 0.750 & 0.080 & 0.760 & 0.025 \\
\midrule
Workflow (160) & Opus 5.5 & 0.925 & 0.097 & -- & -- & -- & -- & -- & -- \\
 & DeepSeek V4.1 Flash & 0.487 & 0.020 & \textbf{0.512} & 0.020 & 0.506 & 0.026 & 0.344 & 0.030 \\
 & GPT-OSS-120B & \textbf{0.094} & 0.015 & 0.019 & 0.003 & 0.019 & 0.004 & -- & -- \\
 & MiniMax M2.5 & \textbf{0.119} & 0.094 & 0.006 & 0.001 & 0.000 & 0.002 & -- & -- \\
 & GLM-5 & \textbf{0.138} & 0.129 & 0.069 & 0.074 & 0.062 & 0.055 & -- & -- \\
 & Qwen3-Next-80B & 0.000 & 0.097 & \textbf{0.019} & 0.078 & \textbf{0.019} & 0.075 & 0.000 & 0.050 \\
\midrule
Terminal (30) & Opus 5.5 & 0.433 & 1.930 & -- & -- & -- & -- & -- & -- \\
 & DeepSeek V4.1 Flash & \textbf{0.100}$^\ddagger$ & 0.307 & \textbf{0.100}$^\ddagger$ & 0.363 & \textbf{0.100} & 0.438 & 0.000 & 0.204 \\
 & GPT-OSS-120B & 0.000 & 0.175 & -- & -- & -- & -- & -- & -- \\
 & MiniMax M2.5 & 0.000 & 0.441 & -- & -- & -- & -- & -- & -- \\
 & GLM-5 & 0.000 & 0.350 & 0.000 & 0.195 & 0.000 & 0.224 & -- & -- \\
 & Qwen3-Next-80B & 0.000 & 0.063 & 0.000 & 0.056 & 0.000 & 0.046 & 0.000 & 0.039 \\
\bottomrule
\end{tabular}
\end{table}

\begin{figure}[t]
\centering
\includegraphics[width=\linewidth]{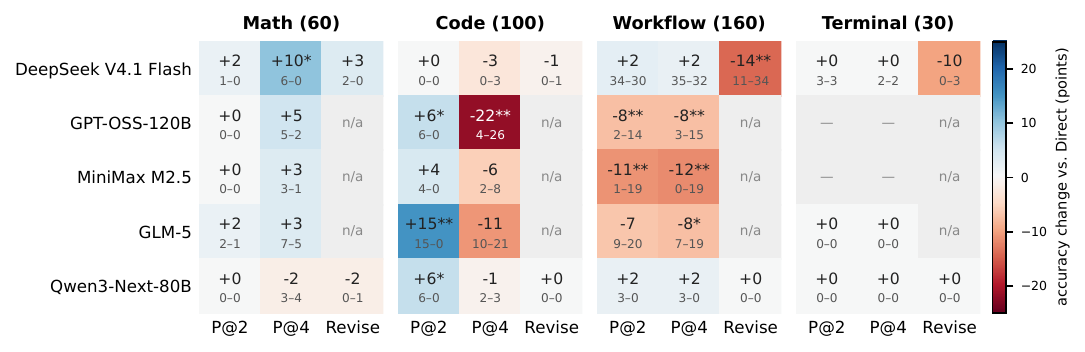}
\caption{Frozen protocol matrix: accuracy change against the same model's Direct run, in percentage points (large) and paired wins--losses (small). Stars: uncorrected exact sign tests ($^{*}p<0.05$, $^{**}p<0.01$); Holm-adjusted tests are in Table~\ref{tab:holm}. n/a: excluded from revision by context; ---: not run.}
\label{fig:delta}
\end{figure}

\begin{table}[t]
\centering\small
\caption{Frozen protocol matrix: mean paired accuracy change $\Delta$ (pp) against Direct over the fixed panel of low-cost models (five for Parallel, three on terminal tasks, two for Revise@4), with pointwise 95\% task-bootstrap intervals and pooled wins--losses (W--L). The intervals describe this panel, not unseen model families.}
\label{tab:pooled}
\setlength{\tabcolsep}{4pt}
\begin{tabular*}{\linewidth}{@{\extracolsep{\fill}}lrcrrcrrcr@{}}
\toprule
 & \multicolumn{3}{c}{Parallel@2} & \multicolumn{3}{c}{Parallel@4} & \multicolumn{3}{c}{Revise@4} \\
\cmidrule(lr){2-4}\cmidrule(lr){5-7}\cmidrule(l){8-10}
Domain & $\Delta$ & 95\% interval & W--L & $\Delta$ & 95\% interval & W--L & $\Delta$ & 95\% interval & W--L \\
\midrule
Math & $+0.7$ & $[-0.7,+2.0]$ & 3--1 & $+4.0$ & $[+0.3,+7.7]$ & 24--12 & $+0.8$ & $[-1.7,+3.3]$ & 2--1 \\
Code & $+6.2$ & $[+4.2,+8.2]$ & 31--0 & $-8.6$ & $[-11.8,-5.4]$ & 18--61 & $-0.5$ & $[-1.5,+0.0]$ & 0--1 \\
Workflow & $-4.2$ & $[-7.2,-1.4]$ & 49--83 & $-4.6$ & $[-7.4,-1.9]$ & 48--85 & $-7.2$ & $[-11.2,-3.4]$ & 11--34 \\
Terminal & $+0.0$ & $[-5.6,+5.6]$ & 3--3 & $+0.0$ & $[-4.4,+4.4]$ & 2--2 & $-5.0$ & $[-10.0,+0.0]$ & 0--3 \\
\bottomrule
\end{tabular*}
\end{table}

Table~\ref{tab:summary} reports the accuracy and mean cost of every configuration.
Figure~\ref{fig:delta} gives each model's paired change against its own Direct run, and Table~\ref{tab:pooled} gives the panel means.
A larger allowance does not order the protocols by accuracy.
On programming, Parallel@2 gains 31 task--model outcomes and loses none, improving four models while Flash stays at 99/100.
Parallel@4, which adds a third candidate and a same-model judge, records 18 wins and 61 losses, a panel mean of $-8.6$~pp.

Mathematics responds differently.
Parallel@4 raises the accuracy of four of five models, with a panel mean of $+4.0$~pp and a 95\% interval of $[0.3,7.7]$.
Flash improves from 46 to 52 of 60 tasks (nominal $p=0.031$, not significant after Holm correction), while Qwen3-Next moves from 31 to 30.
Two-way voting preserves the first parsed answer, so its small mathematics change does not measure the value of self-consistency at larger sample counts.

No agent protocol raises the panel mean.
On workflows, plan-level parallelism lowers it by 4.2 and 4.6~pp for Parallel@2 and Parallel@4: Flash and Qwen3-Next gain a few tasks, while GLM-5, GPT-OSS, and MiniMax lose.
Periodic self-review lowers Flash from 78 to 55 of 160 workflows, with 11 wins and 34 losses (Holm-adjusted $p=0.033$).
On terminal tasks, Parallel preserves aggregate accuracy, although Flash exchanges solved tasks with Direct; Revise lowers Flash from 3/30 to 0/30.

\begin{finding}{1}{Gains depend on both the domain and the choice rule.}
Public-example selection improves four coding models and preserves the fifth, whereas joint selection lowers the programming score; this contrast motivates separating candidate generation from selection and submission.
Because feedback and difficulty change together across domains, cross-domain differences do not isolate either as a cause.
\end{finding}

\subsection{RQ2: Converting candidate coverage into task success}
\label{sec:results-rq2}

\begin{figure}[t]
\centering
\includegraphics[width=\linewidth]{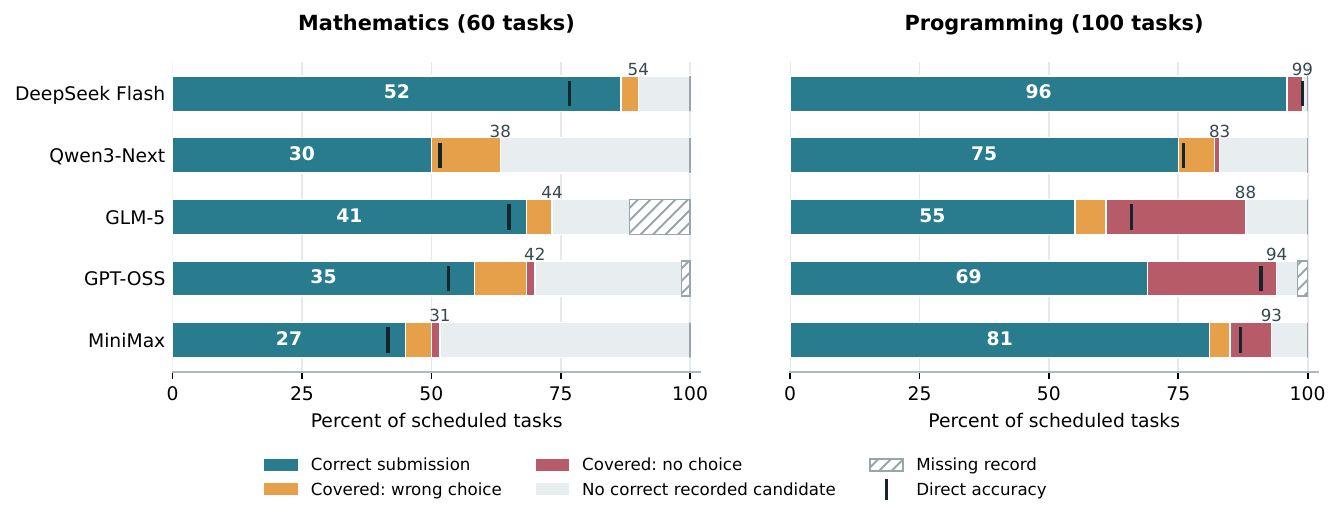}
\caption{Candidate coverage and conversion under frozen Parallel@4. Each bar partitions scheduled tasks. White numbers count correct submissions, labels above bars count covered tasks, and black ticks mark Direct accuracy. ``No choice'' denotes no valid index; ``wrong choice'' includes selecting an empty candidate. Same-pool alternatives are compared in Figure~\ref{fig:selector-replay-cost} and Appendix~\ref{app:selector-replay}.}
\label{fig:conversion}
\end{figure}

\begin{figure}[t]
\centering
\includegraphics[width=0.82\linewidth]{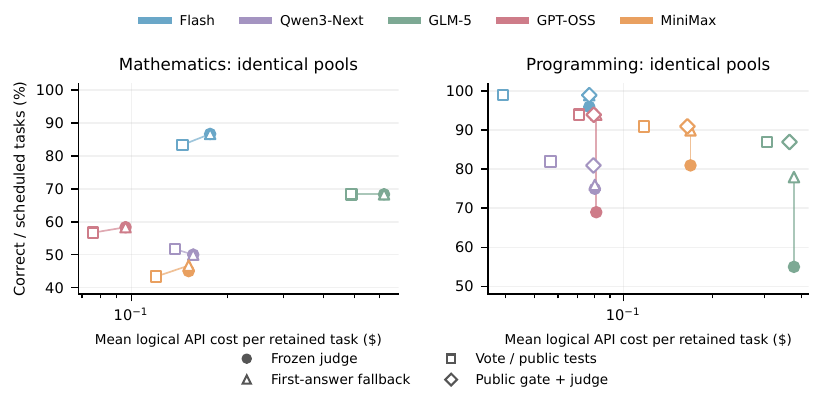}
\caption{Same-pool choices and logical API cost. Accuracy uses scheduled denominators (60 mathematics or 100 coding tasks per model), retaining missing records as failures; cost averages over retained records. Mathematics voting and programming public selection remove judge-stage calls. Fallback retains already-spent cost. Gate costs assume calling the unchanged judge only on public-score ties (Appendix~\ref{app:selector-replay}). Local execution expense is reported separately.}
\label{fig:selector-replay-cost}
\end{figure}

\paragraph{Coverage does not guarantee a correct submission.}
Figure~\ref{fig:conversion} separates generation from conversion in Parallel@4.
In mathematics, Qwen3-Next generates a correct candidate for 38 tasks but submits only 30 correctly, and each lost opportunity is a wrong choice.
In programming, GLM-5 covers 88 tasks but submits 55 correctly; missing choices account for most of the gap (27 of the 33 lost pools).
In these cases the additional computation has already produced useful answers, and the remaining question is how to deliver them.

\paragraph{Selection and fallback recover much of this potential.}
On the same programming pools, retaining the first nonempty candidate when judging yields no answer raises correct submissions from 376 to 437 of 500 scheduled cells (Figure~\ref{fig:selector-replay-cost}).
This alone reverses the mean change against Direct from $-8.6$ to $+3.6$~pp, with the judge's cost still charged.
Public-example selection yields 453/500 ($+6.8$~pp against Direct), converting 453 of 457 covered pools while saving 12--49\% of logical API cost across models.
Selection also adds value beyond preserving an output: among the 98 pools with both correct and incorrect candidates, it solves 94 versus 78 for fallback (16 paired wins, no losses).
In mathematics, judging with the same fallback yields 186 correct submissions on 292 retained pools, compared with 182 for vote@3.
This smaller observed difference has a paired interval spanning zero, while voting saves 12--21\% of API cost.
Grading sensitivity, per-model counts, uncertainty, and accounting details appear in Appendix~\ref{app:selector-replay}.

\begin{finding}{2}{Candidate coverage becomes useful performance through task-appropriate selection and reliable submission.}
Programming's apparent TTS loss reverses with answer-preserving fallback, and public examples recover nearly all recorded coverage.
The smaller mathematics difference shows why a judge's marginal value must be measured against a cheaper choice rule.
\end{finding}

\paragraph{Revision and protocol execution.}
Revision likewise requires separating content changes from lost submissions.
Flash's two mathematics recoveries comprise one repaired answer and one recovered empty output, and Qwen3-Next changes one correct answer to a wrong one.
Flash's sole coding loss follows a truncated critique and an empty final submission; retaining its completed initial program restores the score at unchanged cost.
Lower revision accuracy alone therefore does not establish that a correct answer was rewritten incorrectly (Appendix~\ref{app:revision-fallback}).
For agents, missing plans or malformed choices frequently prevent the GPT-OSS and MiniMax workflow protocols from reaching execution (Appendix~\ref{app:diagnostics}).
These outcomes measure the model--protocol--interface combination and leave the quality of unexecuted plans unresolved.

\begin{finding}{3}{Agent completion and plan quality remain separate questions.}
Frequent pre-execution failures prevent judging the value of a functioning multi-plan strategy.
Static fallback does not transfer to a changed environment, so agent interface validation and repeated episodes remain necessary.
\end{finding}

\begin{figure}[t]
\centering
\includegraphics[width=\linewidth]{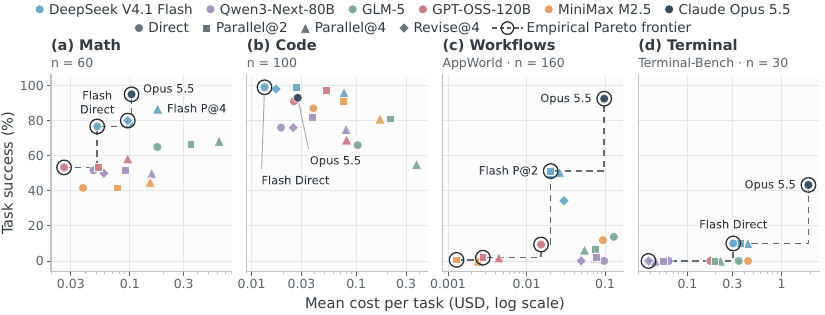}
\caption{Cost and accuracy of the frozen configurations; rings mark their Pareto frontier. Selector replays are compared separately in Figure~\ref{fig:selector-replay-cost}. Costs are mean logical API costs per task over retained cells (DeepSeek peak rates; Bedrock list rates, with Bedrock caching only for Opus agent requests). Code uses the official grades (Opus 93\%, 99\% with \texttt{numpy}), and very low workflow costs can reflect early protocol failures. The frontiers are point estimates, not certified dominance relations.}
\label{fig:pareto}
\end{figure}

\subsection{RQ3: When extra inference is competitive with a stronger model}
\label{sec:results-rq3}

\paragraph{Within-model gains and deployment value are different tests.}
Figure~\ref{fig:pareto} places every frozen configuration by its mean cost and accuracy.
Opus~5.5 Direct leads in mathematics (57/60 against Flash's best 52/60), workflows (148/160 against 82/160), and terminal tasks (13/30 against 3/30).
In programming, GPT-OSS Parallel@2 exceeds the anchor's frozen score (97 against 93) but costs more (\$0.052 against \$0.028), while Flash Direct scores 99/100 at \$0.013.
Supplying the anchor's missing \texttt{numpy} dependency raises it to 99/100 without new inference (Appendix~\ref{app:accounting}), so GPT-OSS's advantage is runtime-dependent.
Flash Direct still dominates the tested coding replays in API cost and observed accuracy.

\paragraph{The mathematics trade-off changes with the price paid.}
At peak rates, Flash judging (52/60, \$0.176/task) and vote@3 (50/60, \$0.144) are both dominated by Opus Direct (57/60, \$0.103).
At half price, they cost \$0.088 and \$0.072 and both lie on the expanded frontier.
The replay thus adds a cheaper operating point, and judging buys two further correct answers (Appendix~\ref{app:selector-replay}).

\begin{finding}{4}{Whether test-time scaling pays off depends on the alternatives and on the price basis.}
Coding replays substantially improve several models without displacing Flash Direct.
Discounted mathematics inference adds accuracy--cost trade-offs while remaining below the stronger reference.
The choice depends on the model, selector, runtime, and price.
\end{finding}

\section{Discussion}
\label{sec:discussion}

\paragraph{Test-time potential is a conversion profile.}
The most informative distinction in our measurements is between producing a correct candidate and successfully submitting it.
Qwen3-Next's mathematics coverage exceeds its Direct accuracy while judging does not.
GLM-5's 88 covered programming pools produce 55 correct submissions under the original judge, 78 with nonempty fallback, and 87 with public-example selection.
An aggregate gain alone would hide these opportunities, and an aggregate loss alone would conflate wrong decisions with stages that return no decision.
Separating coverage, conversion, and revision transitions by output availability makes these cases distinguishable without proposing a new composite score.
The contribution is therefore a concrete measurement framework and its cross-domain evidence, rather than a claim that one protocol realizes all of a model's potential.

\paragraph{Improve the limiting stage before increasing its allowance.}
The coding results motivate making stage completion an explicit design objective: most losses against Direct occur without a usable final selection.
Returning a completed candidate when judging fails recovers 61 coding submissions without new inference, which exposes an avoidable failure-handling loss rather than a limit of candidate generation.
For models such as Qwen3-Next in mathematics, however, the selection stage completes and still discards correct candidates, so the relevant target is decision quality.
This distinction sharpens the connection to work separating sampling from selection~\citep{brown2024monkeys} and examining feedback in self-correction~\citep{huang2024cannot,kamoi2024survey}.

\paragraph{Measure a judge against a cheaper choice rule.}
On programming, public-example selection converts 453 of the 457 covered pools, improving on the frozen judge while avoiding its API cost, and it chooses better within pools of mixed correctness rather than merely retaining an answer (Appendix~\ref{app:mixed-pools}).
A public-first gate that consults the judge only on public-score ties still calls it on 465 of 498 retained pools, because scores often tie, and yields 452 correct submissions.
Public evidence is thus useful even when incomplete, and in these pools the extra judge does not improve on the earliest-best tie rule; this is evidence about the tested selector, not about all possible judges or learned verifiers.
On mathematics, judging yields three more correct submissions than voting on identical pools, or four with nonempty fallback, and both paired intervals cross zero, while removing the judge saves 12--21\% of the judge-based configuration's logical API cost.
The saved Compass judgments preserve the direction of the difference, with 194 correct submissions for judging with fallback versus 189 for voting (Appendix~\ref{app:grading-sensitivity}), but neither grader establishes equivalence between the selectors.

\paragraph{Choosing a method is also choosing a model and a runtime.}
A deployment decision should compare an entire configuration---model, protocol, harness, and price---against the available alternatives.
GLM-5's sizeable coding gain is valuable evidence about its response to inference, yet a cheaper Flash Direct configuration solves more tasks.
After incorporating the replays, Flash Direct remains the coding frontier.
In mathematics, the discounted replays add operating points: at half price, Flash vote@3 reaches 50/60 at \$0.072/task and judging 52/60 at \$0.088, whereas at peak rates both are dominated by Opus Direct.
The anchor's coding sensitivity to one missing library further shows why a performance difference should not automatically be interpreted as a capability difference.
These observations connect the scientific question of attainable gains to a practical decision: whether to spend the next unit of budget on another candidate, a more reliable selection mechanism, or a different base model.

\paragraph{What should be reported in a TTS evaluation.}
Our results motivate reporting the Direct operating point and actual resource use, candidate coverage and availability where observable, the outcome of each selection stage, revision transitions, and final native grades alongside termination status.
The budget cap and the measured cost answer different questions, and both should be visible.
A stronger Direct reference makes deployment claims assessable, while explicit runtime and price sensitivity analyses prevent an incidental environment or tariff from becoming a general conclusion.
For agent protocols, unexecuted plans do not provide an oracle coverage estimate; additional controlled executions would be required to separate plan quality from execution reliability.

\section{Related Work}
\label{sec:related}

\begin{table}[!ht]
\centering
\small
\setlength{\tabcolsep}{4.5pt}
\caption{Controlled studies of test-time scaling. M/C/A/O: mathematics, code, agent tasks, other; P/S: parallel, sequential. \emph{Judge charged}: selection calls count against the generation budget (---: no judge call). \emph{vs.\ stronger}: a cheaper model with more compute is compared with a stronger model.}
\label{tab:related}
\begin{tabular*}{\linewidth}{@{\extracolsep{\fill}}lcccccc@{}}
\toprule
Study & Domains & Methods & \shortstack{Judge\\charged} & Budget unit & \shortstack{Reasoning\\models} & vs.\ stronger \\
\midrule
\citet{snell2025scaling}          & M     & P, S & \ding{55} & samples, FLOPs & \ding{55} & \ding{51} \\
\citet{wu2025inference}           & M, C  & P    & \ding{55} & FLOPs          & \ding{55} & \ding{51} \\
\citet{wang2024tokeneconomies}    & M, O  & P, S & \ding{51} & tokens, \$     & \ding{55} & \ding{55} \\
\citet{chen2024morecalls}         & O     & P    & \ding{55} & calls          & \ding{55} & \ding{55} \\
\citet{singhi2025solve}           & M, O  & P    & \ding{51} & FLOPs          & \ding{51} & \ding{55} \\
\citet{ghosal2025thinking}        & M     & P, S & ---       & tokens         & \ding{51} & \ding{55} \\
\citet{zeng2025revisiting}        & M, O  & P, S & ---       & tokens         & \ding{51} & \ding{55} \\
\citet{stroebl2026limits}         & C     & P    & ---       & samples        & \ding{55} & \ding{51} \\
\citet{kapoor2026hal}             & A     & S    & ---       & \$             & \ding{51} & partly \\
\citet{li2026agentbench}          & A, M  & P, S & \ding{55} & samples, tokens & \ding{51} & \ding{55} \\
\citet{balachandran2025inference} & M, O  & P, S & \ding{55} & tokens, calls  & \ding{51} & \ding{51} \\
\midrule
\method (ours)                   & M, C, A & P, S & \ding{51} & tokens, \$    & \ding{51} & \ding{51} \\
\bottomrule
\end{tabular*}
\end{table}

\paragraph{Test-time scaling.}
Parallel methods generate several candidates and select one, with a trained verifier that scores answers or reasoning steps~\citep{cobbe2021verifiers,lightman2024verify,zhang2025genrm,saadfalcon2025weaver} or without one, by majority vote~\citep{wang2023selfconsistency,chen2023usc}, by executing programs on tests that the model writes~\citep{chen2023codet}, or by letting the model judge the candidates~\citep{chen2025knockout,zhao2025sample}.
Selection often limits the gains: coverage keeps rising with more samples while voting and reward-model selection level off~\citep{brown2024monkeys}, an imperfect verifier caps the accuracy that resampling can reach~\citep{stroebl2026limits}, and verification draws on the generation budget~\citep{singhi2025solve}.
Sequential methods revise an earlier attempt~\citep{madaan2023selfrefine}.
Without external feedback, self-correction rarely improves reasoning and can turn correct answers into wrong ones~\citep{huang2024cannot,stechly2025selfverification,kamoi2024survey}, whereas feedback from tools, execution, or the environment~\citep{gou2024critic,chen2024selfdebug,shinn2023reflexion} and a model trained to revise~\citep{snell2025scaling} help more, at a cost that can offset the gain~\citep{olausson2024selfrepair}.
Reasoning models also check their work inside one long chain of thought~\citep{muennighoff2025s1}, which beyond a point does worse than parallel sampling at the same budget~\citep{ghosal2025thinking,zeng2025revisiting}, and tree search~\citep{yao2023tot} and debate~\citep{du2024debate} combine the two families.
\method runs the simplest verifier-free form of each family that a reasoning model can execute by itself.

\paragraph{Controlled comparisons.}
Several studies compare test-time scaling methods under a matched budget (Table~\ref{tab:related}).
In mathematics, a smaller model given more inference compute can match or surpass a larger one on questions it already solves at a non-trivial rate, and voting saturates as samples grow~\citep{snell2025scaling,wu2025inference}.
Across single-turn reasoning tasks, self-consistency frequently outperforms more elaborate strategies at equal budget~\citep{wang2024tokeneconomies}, and accuracy can first rise and then fall as the number of calls grows~\citep{chen2024morecalls}.
For agents, higher reasoning effort gives equal or lower accuracy in 21 of 36 model--agent--benchmark combinations~\citep{kapoor2026hal}, and neither sequential nor parallel scaling helps general agents much~\citep{li2026agentbench}.
A concurrent report covers mathematics, science questions, and planning across nine models~\citep{balachandran2025inference}.
None of these studies combines mathematics, programming, and agent tasks with charged selection calls and a comparison against a stronger model, which is the combination \method provides.

\paragraph{Cost-aware evaluation.}
Holistic evaluation suites report efficiency in runtime and hardware units~\citep{liang2023helm}, agent evaluations argue for and adopt joint reporting of dollar cost and accuracy~\citep{kapoor2025agentsmatter,kapoor2026hal}, and cost-of-pass measures the expected dollar cost of one correct answer~\citep{erol2026costofpass}.
Test-time scaling studies mostly count FLOPs~\citep{snell2025scaling,wu2025inference}, yet the unit of cost matters: inference demand changes the compute-optimal model size~\citep{sardana2024chinchilla}, memory-access cost removes much of the apparent advantage of small models under best-of-$N$ sampling and long reasoning~\citep{sadhukhan2025kinetics}, and models often reason far longer than a problem requires~\citep{chen2025overthinking,han2025tale}.
Cascades and routers instead choose a model for each query~\citep{chen2024frugalgpt,yue2024cascades,gupta2024cascades,aggarwal2024automix,ding2024hybrid,ong2025routellm,dekoninck2025cascade}, sometimes together with the number of samples~\citep{ding2025bestroute}, and adaptive stopping rules reduce the samples a single model needs~\citep{aggarwal2023adaptive,li2024esc,damani2025learning}.
These approaches compare models at their native settings or assume that sampling a cheaper model is the more economical path, an assumption that \method tests directly.

\section{Conclusion}
\label{sec:conclusion}

We introduced \method, a benchmark and evaluation framework that measures how additional inference becomes task success, and at what cost, across 350 sealed tasks in mathematics, programming, and agentic environments.
A recurring lesson is that the most informative distinction lies between producing a correct candidate and successfully submitting it.
On fixed programming pools, public-example selection raises correct submissions from 376 to 453, and output-preserving fallback alone recovers 61, so selection and failure handling can reverse an observed TTS loss.
In mathematics, judging with fallback adds four correct submissions over cheaper voting, and discounted vote@3 adds a deployment trade-off, while Flash Direct retains the coding frontier.
For agents, interfaces require validation before their outcomes can be attributed to planning.

Together, these measurements connect candidate coverage, executable selection rules, and cost to the practical question of inference allocation: whether the next unit of budget is better spent on another candidate, a more reliable selection mechanism, or a different base model.

\bibliographystyle{iclr2027_conference}
\bibliography{main}

@inproceedings{wang2023selfconsistency,
  author       = {Xuezhi Wang and
                  Jason Wei and
                  Dale Schuurmans and
                  Quoc V. Le and
                  Ed H. Chi and
                  Sharan Narang and
                  Aakanksha Chowdhery and
                  Denny Zhou},
  title        = {Self-Consistency Improves Chain of Thought Reasoning in Language Models},
  booktitle    = {The Eleventh International Conference on Learning Representations,
                  {ICLR} 2023, Kigali, Rwanda, May 1-5, 2023},
  publisher    = {OpenReview.net},
  year         = {2023},
  url          = {https://openreview.net/forum?id=1PL1NIMMrw},
  bibsource    = {dblp computer science bibliography, https://dblp.org}
}

@article{chen2023usc,
  author       = {Xinyun Chen and
                  Renat Aksitov and
                  Uri Alon and
                  Jie Ren and
                  Kefan Xiao and
                  Pengcheng Yin and
                  Sushant Prakash and
                  Charles Sutton and
                  Xuezhi Wang and
                  Denny Zhou},
  title        = {Universal Self-Consistency for Large Language Model Generation},
  journal      = {CoRR},
  volume       = {abs/2311.17311},
  year         = {2023},
  url          = {https://doi.org/10.48550/arXiv.2311.17311},
  doi          = {10.48550/ARXIV.2311.17311},
  eprinttype   = {arXiv},
  eprint       = {2311.17311},
  bibsource    = {dblp computer science bibliography, https://dblp.org}
}

@article{cobbe2021verifiers,
  author       = {Karl Cobbe and
                  Vineet Kosaraju and
                  Mohammad Bavarian and
                  Mark Chen and
                  Heewoo Jun and
                  Lukasz Kaiser and
                  Matthias Plappert and
                  Jerry Tworek and
                  Jacob Hilton and
                  Reiichiro Nakano and
                  Christopher Hesse and
                  John Schulman},
  title        = {Training Verifiers to Solve Math Word Problems},
  journal      = {CoRR},
  volume       = {abs/2110.14168},
  year         = {2021},
  url          = {https://arxiv.org/abs/2110.14168},
  eprinttype   = {arXiv},
  eprint       = {2110.14168},
  bibsource    = {dblp computer science bibliography, https://dblp.org}
}

@inproceedings{lightman2024verify,
  author       = {Hunter Lightman and
                  Vineet Kosaraju and
                  Yuri Burda and
                  Harrison Edwards and
                  Bowen Baker and
                  Teddy Lee and
                  Jan Leike and
                  John Schulman and
                  Ilya Sutskever and
                  Karl Cobbe},
  title        = {Let's Verify Step by Step},
  booktitle    = {The Twelfth International Conference on Learning Representations,
                  {ICLR} 2024, Vienna, Austria, May 7-11, 2024},
  publisher    = {OpenReview.net},
  year         = {2024},
  url          = {https://openreview.net/forum?id=v8L0pN6EOi},
  bibsource    = {dblp computer science bibliography, https://dblp.org}
}

@inproceedings{zhang2025genrm,
  author       = {Lunjun Zhang and
                  Arian Hosseini and
                  Hritik Bansal and
                  Mehran Kazemi and
                  Aviral Kumar and
                  Rishabh Agarwal},
  title        = {Generative Verifiers: Reward Modeling as Next-Token Prediction},
  booktitle    = {The Thirteenth International Conference on Learning Representations,
                  {ICLR} 2025, Singapore, April 24-28, 2025},
  publisher    = {OpenReview.net},
  year         = {2025},
  url          = {https://openreview.net/forum?id=Ccwp4tFEtE},
  bibsource    = {dblp computer science bibliography, https://dblp.org}
}

@inproceedings{zhao2025sample,
  author       = {Eric Zhao and
                  Pranjal Awasthi and
                  Sreenivas Gollapudi},
  editor       = {Aarti Singh and
                  Maryam Fazel and
                  Daniel Hsu and
                  Simon Lacoste{-}Julien and
                  Felix Berkenkamp and
                  Tegan Maharaj and
                  Kiri Wagstaff and
                  Jerry Zhu},
  title        = {Sample, Scrutinize and Scale: Effective Inference-Time Search by Scaling
                  Verification},
  booktitle    = {Forty-second International Conference on Machine Learning, {ICML}
                  2025, Vancouver, BC, Canada, July 13-19, 2025},
  series       = {Proceedings of Machine Learning Research},
  volume       = {267},
  publisher    = {{PMLR} / OpenReview.net},
  year         = {2025},
  url          = {https://proceedings.mlr.press/v267/zhao25a.html},
  bibsource    = {dblp computer science bibliography, https://dblp.org}
}

@inproceedings{chen2025knockout,
  author       = {Yanxi Chen and
                  Xuchen Pan and
                  Yaliang Li and
                  Bolin Ding and
                  Jingren Zhou},
  editor       = {Danielle Belgrave and
                  Cheng Zhang and
                  Laura N. Montoya and
                  Hsuan{-}Tien Lin and
                  Razvan Pascanu and
                  Piotr Koniusz and
                  Marzyeh Ghassemi and
                  Nancy Chen and
                  Iv{\'{a}}n Vladimir Meza Ru{\'{\i}}z and
                  Arturo Loaiza{-}Bonilla},
  title        = {Provable Scaling Laws for the Test-Time Compute of Large Language
                  Models},
  booktitle    = {Advances in Neural Information Processing Systems 38: Annual Conference
                  on Neural Information Processing Systems 2025, NeurIPS 2025, San Diego,
                  CA, USA, December 2-7, 2025 / Mexico City, Mexico, November 30 - December
                  5, 2025},
  year         = {2025},
  url          = {http://papers.nips.cc/paper\_files/paper/2025/hash/f3f607e4c13bd1cb8885de44b4ec45b7-Abstract-Conference.html},
  bibsource    = {dblp computer science bibliography, https://dblp.org}
}

@inproceedings{chen2023codet,
  author       = {Bei Chen and
                  Fengji Zhang and
                  Anh Nguyen and
                  Daoguang Zan and
                  Zeqi Lin and
                  Jian{-}Guang Lou and
                  Weizhu Chen},
  title        = {CodeT: Code Generation with Generated Tests},
  booktitle    = {The Eleventh International Conference on Learning Representations,
                  {ICLR} 2023, Kigali, Rwanda, May 1-5, 2023},
  publisher    = {OpenReview.net},
  year         = {2023},
  url          = {https://openreview.net/forum?id=ktrw68Cmu9c},
  bibsource    = {dblp computer science bibliography, https://dblp.org}
}

@inproceedings{madaan2023selfrefine,
  author       = {Aman Madaan and
                  Niket Tandon and
                  Prakhar Gupta and
                  Skyler Hallinan and
                  Luyu Gao and
                  Sarah Wiegreffe and
                  Uri Alon and
                  Nouha Dziri and
                  Shrimai Prabhumoye and
                  Yiming Yang and
                  Shashank Gupta and
                  Bodhisattwa Prasad Majumder and
                  Katherine Hermann and
                  Sean Welleck and
                  Amir Yazdanbakhsh and
                  Peter Clark},
  editor       = {Alice Oh and
                  Tristan Naumann and
                  Amir Globerson and
                  Kate Saenko and
                  Moritz Hardt and
                  Sergey Levine},
  title        = {Self-Refine: Iterative Refinement with Self-Feedback},
  booktitle    = {Advances in Neural Information Processing Systems 36: Annual Conference
                  on Neural Information Processing Systems 2023, NeurIPS 2023, New Orleans,
                  LA, USA, December 10 - 16, 2023},
  year         = {2023},
  url          = {http://papers.nips.cc/paper\_files/paper/2023/hash/91edff07232fb1b55a505a9e9f6c0ff3-Abstract-Conference.html},
  bibsource    = {dblp computer science bibliography, https://dblp.org}
}

@inproceedings{gou2024critic,
  author       = {Zhibin Gou and
                  Zhihong Shao and
                  Yeyun Gong and
                  Yelong Shen and
                  Yujiu Yang and
                  Nan Duan and
                  Weizhu Chen},
  title        = {{CRITIC:} Large Language Models Can Self-Correct with Tool-Interactive
                  Critiquing},
  booktitle    = {The Twelfth International Conference on Learning Representations,
                  {ICLR} 2024, Vienna, Austria, May 7-11, 2024},
  publisher    = {OpenReview.net},
  year         = {2024},
  url          = {https://openreview.net/forum?id=Sx038qxjek},
  bibsource    = {dblp computer science bibliography, https://dblp.org}
}

@inproceedings{chen2024selfdebug,
  author       = {Xinyun Chen and
                  Maxwell Lin and
                  Nathanael Sch{\"{a}}rli and
                  Denny Zhou},
  title        = {Teaching Large Language Models to Self-Debug},
  booktitle    = {The Twelfth International Conference on Learning Representations,
                  {ICLR} 2024, Vienna, Austria, May 7-11, 2024},
  publisher    = {OpenReview.net},
  year         = {2024},
  url          = {https://openreview.net/forum?id=KuPixIqPiq},
  bibsource    = {dblp computer science bibliography, https://dblp.org}
}

@inproceedings{shinn2023reflexion,
  author       = {Noah Shinn and
                  Federico Cassano and
                  Ashwin Gopinath and
                  Karthik Narasimhan and
                  Shunyu Yao},
  editor       = {Alice Oh and
                  Tristan Naumann and
                  Amir Globerson and
                  Kate Saenko and
                  Moritz Hardt and
                  Sergey Levine},
  title        = {Reflexion: language agents with verbal reinforcement learning},
  booktitle    = {Advances in Neural Information Processing Systems 36: Annual Conference
                  on Neural Information Processing Systems 2023, NeurIPS 2023, New Orleans,
                  LA, USA, December 10 - 16, 2023},
  year         = {2023},
  url          = {http://papers.nips.cc/paper\_files/paper/2023/hash/1b44b878bb782e6954cd888628510e90-Abstract-Conference.html},
  bibsource    = {dblp computer science bibliography, https://dblp.org}
}

@inproceedings{muennighoff2025s1,
  author       = {Niklas Muennighoff and
                  Zitong Yang and
                  Weijia Shi and
                  Xiang Lisa Li and
                  Li Fei{-}Fei and
                  Hannaneh Hajishirzi and
                  Luke Zettlemoyer and
                  Percy Liang and
                  Emmanuel J. Cand{\`{e}}s and
                  Tatsunori Hashimoto},
  editor       = {Christos Christodoulopoulos and
                  Tanmoy Chakraborty and
                  Carolyn Rose and
                  Violet Peng},
  title        = {s1: Simple test-time scaling},
  booktitle    = {Proceedings of the 2025 Conference on Empirical Methods in Natural
                  Language Processing, {EMNLP} 2025, Suzhou, China, November 4-9, 2025},
  pages        = {20275--20321},
  publisher    = {Association for Computational Linguistics},
  year         = {2025},
  url          = {https://doi.org/10.18653/v1/2025.emnlp-main.1025},
  doi          = {10.18653/V1/2025.EMNLP-MAIN.1025},
  bibsource    = {dblp computer science bibliography, https://dblp.org}
}

@inproceedings{yao2023tot,
  author       = {Shunyu Yao and
                  Dian Yu and
                  Jeffrey Zhao and
                  Izhak Shafran and
                  Tom Griffiths and
                  Yuan Cao and
                  Karthik Narasimhan},
  editor       = {Alice Oh and
                  Tristan Naumann and
                  Amir Globerson and
                  Kate Saenko and
                  Moritz Hardt and
                  Sergey Levine},
  title        = {Tree of Thoughts: Deliberate Problem Solving with Large Language Models},
  booktitle    = {Advances in Neural Information Processing Systems 36: Annual Conference
                  on Neural Information Processing Systems 2023, NeurIPS 2023, New Orleans,
                  LA, USA, December 10 - 16, 2023},
  year         = {2023},
  url          = {http://papers.nips.cc/paper\_files/paper/2023/hash/271db9922b8d1f4dd7aaef84ed5ac703-Abstract-Conference.html},
  bibsource    = {dblp computer science bibliography, https://dblp.org}
}

@inproceedings{du2024debate,
  author       = {Yilun Du and
                  Shuang Li and
                  Antonio Torralba and
                  Joshua B. Tenenbaum and
                  Igor Mordatch},
  editor       = {Ruslan Salakhutdinov and
                  Zico Kolter and
                  Katherine A. Heller and
                  Adrian Weller and
                  Nuria Oliver and
                  Jonathan Scarlett and
                  Felix Berkenkamp},
  title        = {Improving Factuality and Reasoning in Language Models through Multiagent
                  Debate},
  booktitle    = {Forty-first International Conference on Machine Learning, {ICML} 2024,
                  Vienna, Austria, July 21-27, 2024},
  series       = {Proceedings of Machine Learning Research},
  volume       = {235},
  pages        = {11733--11763},
  publisher    = {{PMLR} / OpenReview.net},
  year         = {2024},
  url          = {https://proceedings.mlr.press/v235/du24e.html},
  bibsource    = {dblp computer science bibliography, https://dblp.org}
}

@inproceedings{snell2025scaling,
  author       = {Charlie Victor Snell and
                  Jaehoon Lee and
                  Kelvin Xu and
                  Aviral Kumar},
  title        = {Scaling {LLM} Test-Time Compute Optimally Can be More Effective than
                  Scaling Parameters for Reasoning},
  booktitle    = {The Thirteenth International Conference on Learning Representations,
                  {ICLR} 2025, Singapore, April 24-28, 2025},
  publisher    = {OpenReview.net},
  year         = {2025},
  url          = {https://openreview.net/forum?id=4FWAwZtd2n},
  bibsource    = {dblp computer science bibliography, https://dblp.org}
}

@inproceedings{damani2025learning,
  author       = {Mehul Damani and
                  Idan Shenfeld and
                  Andi Peng and
                  Andreea Bobu and
                  Jacob Andreas},
  title        = {Learning How Hard to Think: Input-Adaptive Allocation of {LM} Computation},
  booktitle    = {The Thirteenth International Conference on Learning Representations,
                  {ICLR} 2025, Singapore, April 24-28, 2025},
  publisher    = {OpenReview.net},
  year         = {2025},
  url          = {https://openreview.net/forum?id=6qUUgw9bAZ},
  bibsource    = {dblp computer science bibliography, https://dblp.org}
}

@article{brown2024monkeys,
  author       = {Bradley C. A. Brown and
                  Jordan Juravsky and
                  Ryan Ehrlich and
                  Ronald Clark and
                  Quoc V. Le and
                  Christopher R{\'{e}} and
                  Azalia Mirhoseini},
  title        = {Large Language Monkeys: Scaling Inference Compute with Repeated Sampling},
  journal      = {CoRR},
  volume       = {abs/2407.21787},
  year         = {2024},
  url          = {https://doi.org/10.48550/arXiv.2407.21787},
  doi          = {10.48550/ARXIV.2407.21787},
  eprinttype   = {arXiv},
  eprint       = {2407.21787},
  bibsource    = {dblp computer science bibliography, https://dblp.org}
}

@inproceedings{saadfalcon2025weaver,
  author       = {Jon Saad{-}Falcon and
                  Estefany Kelly Buchanan and
                  Mayee F. Chen and
                  Tzu{-}Heng Huang and
                  Brendan McLaughlin and
                  Tanvir Bhathal and
                  Shang Zhu and
                  Ben Athiwaratkun and
                  Frederic Sala and
                  Scott W. Linderman and
                  Azalia Mirhoseini and
                  Christopher R{\'{e}}},
  editor       = {Danielle Belgrave and
                  Cheng Zhang and
                  Laura N. Montoya and
                  Hsuan{-}Tien Lin and
                  Razvan Pascanu and
                  Piotr Koniusz and
                  Marzyeh Ghassemi and
                  Nancy Chen and
                  Iv{\'{a}}n Vladimir Meza Ru{\'{\i}}z and
                  Arturo Loaiza{-}Bonilla},
  title        = {Weaver: Shrinking the Generation-Verification Gap by Scaling Compute
                  for Verification},
  booktitle    = {Advances in Neural Information Processing Systems 38: Annual Conference
                  on Neural Information Processing Systems 2025, NeurIPS 2025, San Diego,
                  CA, USA, December 2-7, 2025 / Mexico City, Mexico, November 30 - December
                  5, 2025},
  year         = {2025},
  url          = {http://papers.nips.cc/paper\_files/paper/2025/hash/715f48023c26ad210321a4c0da67f27c-Abstract-Conference.html},
  bibsource    = {dblp computer science bibliography, https://dblp.org}
}

@inproceedings{stroebl2026limits,
title={The Limits of Inference Scaling Through Resampling},
author={Benedikt Stroebl and Sayash Kapoor and Arvind Narayanan},
booktitle={The Fourteenth International Conference on Learning Representations},
year={2026},
url={https://openreview.net/forum?id=j8H84v6AZ1}
}

@inproceedings{huang2024cannot,
  author       = {Jie Huang and
                  Xinyun Chen and
                  Swaroop Mishra and
                  Huaixiu Steven Zheng and
                  Adams Wei Yu and
                  Xinying Song and
                  Denny Zhou},
  title        = {Large Language Models Cannot Self-Correct Reasoning Yet},
  booktitle    = {The Twelfth International Conference on Learning Representations,
                  {ICLR} 2024, Vienna, Austria, May 7-11, 2024},
  publisher    = {OpenReview.net},
  year         = {2024},
  url          = {https://openreview.net/forum?id=IkmD3fKBPQ},
  bibsource    = {dblp computer science bibliography, https://dblp.org}
}

@article{kamoi2024survey,
  author       = {Ryo Kamoi and
                  Yusen Zhang and
                  Nan Zhang and
                  Jiawei Han and
                  Rui Zhang},
  title        = {When Can LLMs \emph{Actually} Correct Their Own Mistakes? {A} Critical
                  Survey of Self-Correction of LLMs},
  journal      = {Trans. Assoc. Comput. Linguistics},
  volume       = {12},
  pages        = {1417--1440},
  year         = {2024},
  url          = {https://doi.org/10.1162/tacl\_a\_00713},
  doi          = {10.1162/TACL\_A\_00713},
  bibsource    = {dblp computer science bibliography, https://dblp.org}
}

@inproceedings{stechly2025selfverification,
  author       = {Kaya Stechly and
                  Karthik Valmeekam and
                  Subbarao Kambhampati},
  title        = {On the self-verification limitations of large language models on reasoning
                  and planning tasks},
  booktitle    = {The Thirteenth International Conference on Learning Representations,
                  {ICLR} 2025, Singapore, April 24-28, 2025},
  publisher    = {OpenReview.net},
  year         = {2025},
  url          = {https://openreview.net/forum?id=4O0v4s3IzY},
  bibsource    = {dblp computer science bibliography, https://dblp.org}
}

@article{singhi2025solve,
  author       = {Nishad Singhi and
                  Hritik Bansal and
                  Arian Hosseini and
                  Aditya Grover and
                  Kai{-}Wei Chang and
                  Marcus Rohrbach and
                  Anna Rohrbach},
  title        = {When To Solve, When To Verify: Compute-Optimal Problem Solving and
                  Generative Verification for {LLM} Reasoning},
  journal      = {CoRR},
  volume       = {abs/2504.01005},
  year         = {2025},
  url          = {https://doi.org/10.48550/arXiv.2504.01005},
  doi          = {10.48550/ARXIV.2504.01005},
  eprinttype   = {arXiv},
  eprint       = {2504.01005},
  bibsource    = {dblp computer science bibliography, https://dblp.org}
}

@inproceedings{wu2025inference,
  author       = {Yangzhen Wu and
                  Zhiqing Sun and
                  Shanda Li and
                  Sean Welleck and
                  Yiming Yang},
  title        = {Inference Scaling Laws: An Empirical Analysis of Compute-Optimal Inference
                  for {LLM} Problem-Solving},
  booktitle    = {The Thirteenth International Conference on Learning Representations,
                  {ICLR} 2025, Singapore, April 24-28, 2025},
  publisher    = {OpenReview.net},
  year         = {2025},
  url          = {https://openreview.net/forum?id=VNckp7JEHn},
  bibsource    = {dblp computer science bibliography, https://dblp.org}
}

@inproceedings{ghosal2025thinking,
  author       = {Soumya Suvra Ghosal and
                  Souradip Chakraborty and
                  Avinash Reddy and
                  Yifu Lu and
                  Mengdi Wang and
                  Dinesh Manocha and
                  Furong Huang and
                  Mohammad Ghavamzadeh and
                  Amrit Singh Bedi},
  editor       = {Danielle Belgrave and
                  Cheng Zhang and
                  Laura N. Montoya and
                  Hsuan{-}Tien Lin and
                  Razvan Pascanu and
                  Piotr Koniusz and
                  Marzyeh Ghassemi and
                  Nancy Chen and
                  Iv{\'{a}}n Vladimir Meza Ru{\'{\i}}z and
                  Arturo Loaiza{-}Bonilla},
  title        = {Does Thinking More Always Help? Mirage of Test-Time Scaling in Reasoning
                  Models},
  booktitle    = {Advances in Neural Information Processing Systems 38: Annual Conference
                  on Neural Information Processing Systems 2025, NeurIPS 2025, San Diego,
                  CA, USA, December 2-7, 2025 / Mexico City, Mexico, November 30 - December
                  5, 2025},
  year         = {2025},
  url          = {http://papers.nips.cc/paper\_files/paper/2025/hash/fc067ac218430c409d6f65403328f740-Abstract-Conference.html},
  bibsource    = {dblp computer science bibliography, https://dblp.org}
}

@inproceedings{zeng2025revisiting,
  author       = {Zhiyuan Zeng and
                  Qinyuan Cheng and
                  Zhangyue Yin and
                  Yunhua Zhou and
                  Xipeng Qiu},
  editor       = {Wanxiang Che and
                  Joyce Nabende and
                  Ekaterina Shutova and
                  Mohammad Taher Pilehvar},
  title        = {Revisiting the Test-Time Scaling of o1-like Models: Do they Truly
                  Possess Test-Time Scaling Capabilities?},
  booktitle    = {Proceedings of the 63rd Annual Meeting of the Association for Computational
                  Linguistics (Volume 1: Long Papers), {ACL} 2025, Vienna, Austria,
                  July 27 - August 1, 2025},
  pages        = {4651--4665},
  publisher    = {Association for Computational Linguistics},
  year         = {2025},
  url          = {https://doi.org/10.18653/v1/2025.acl-long.232},
  doi          = {10.18653/V1/2025.ACL-LONG.232},
  bibsource    = {dblp computer science bibliography, https://dblp.org}
}

@inproceedings{olausson2024selfrepair,
  author       = {Theo X. Olausson and
                  Jeevana Priya Inala and
                  Chenglong Wang and
                  Jianfeng Gao and
                  Armando Solar{-}Lezama},
  title        = {Is Self-Repair a Silver Bullet for Code Generation?},
  booktitle    = {The Twelfth International Conference on Learning Representations,
                  {ICLR} 2024, Vienna, Austria, May 7-11, 2024},
  publisher    = {OpenReview.net},
  year         = {2024},
  url          = {https://openreview.net/forum?id=y0GJXRungR},
  bibsource    = {dblp computer science bibliography, https://dblp.org}
}

@inproceedings{chen2024morecalls,
  author       = {Lingjiao Chen and
                  Jared Quincy Davis and
                  Boris Hanin and
                  Peter Bailis and
                  Ion Stoica and
                  Matei A. Zaharia and
                  James Y. Zou},
  editor       = {Amir Globersons and
                  Lester Mackey and
                  Danielle Belgrave and
                  Angela Fan and
                  Ulrich Paquet and
                  Jakub M. Tomczak and
                  Cheng Zhang},
  title        = {Are More {LLM} Calls All You Need? Towards the Scaling Properties
                  of Compound {AI} Systems},
  booktitle    = {Advances in Neural Information Processing Systems 37: Annual Conference
                  on Neural Information Processing Systems 2024, NeurIPS 2024, Vancouver,
                  BC, Canada, December 10 - 15, 2024},
  year         = {2024},
  url          = {http://papers.nips.cc/paper\_files/paper/2024/hash/51173cf34c5faac9796a47dc2fdd3a71-Abstract-Conference.html},
  bibsource    = {dblp computer science bibliography, https://dblp.org}
}

@inproceedings{wang2024tokeneconomies,
  author       = {Junlin Wang and
                  Siddhartha Jain and
                  Dejiao Zhang and
                  Baishakhi Ray and
                  Varun Kumar and
                  Ben Athiwaratkun},
  editor       = {Yaser Al{-}Onaizan and
                  Mohit Bansal and
                  Yun{-}Nung Chen},
  title        = {Reasoning in Token Economies: Budget-Aware Evaluation of {LLM} Reasoning
                  Strategies},
  booktitle    = {Proceedings of the 2024 Conference on Empirical Methods in Natural
                  Language Processing, {EMNLP} 2024, Miami, FL, USA, November 12-16,
                  2024},
  pages        = {19916--19939},
  publisher    = {Association for Computational Linguistics},
  year         = {2024},
  url          = {https://doi.org/10.18653/v1/2024.emnlp-main.1112},
  doi          = {10.18653/V1/2024.EMNLP-MAIN.1112},
  bibsource    = {dblp computer science bibliography, https://dblp.org}
}

@article{kapoor2026hal,
  author       = {Sayash Kapoor and
                  Benedikt Stroebl and
                  Peter Kirgis and
                  Nitya Nadgir and
                  Zachary S. Siegel and
                  Boyi Wei and
                  Tianci Xue and
                  Ziru Chen and
                  Felix Chen and
                  Saiteja Utpala and
                  Franck Ndzomga and
                  Dheeraj Oruganty and
                  Sophie Luskin and
                  Kangheng Liu and
                  Botao Yu and
                  Amit Arora and
                  Dongyoon Hahm and
                  Harsh Trivedi and
                  Huan Sun and
                  Juyong Lee and
                  Tengjun Jin and
                  Yifan Mai and
                  Yifei Zhou and
                  Yuxuan Zhu and
                  Rishi Bommasani and
                  Daniel Kang and
                  Dawn Song and
                  Peter Henderson and
                  Yu Su and
                  Percy Liang and
                  Arvind Narayanan},
  title        = {Holistic Agent Leaderboard: The Missing Infrastructure for {AI} Agent
                  Evaluation},
  journal      = {CoRR},
  volume       = {abs/2510.11977},
  year         = {2025},
  url          = {https://doi.org/10.48550/arXiv.2510.11977},
  doi          = {10.48550/ARXIV.2510.11977},
  eprinttype   = {arXiv},
  eprint       = {2510.11977},
  bibsource    = {dblp computer science bibliography, https://dblp.org}
}

@article{li2026agentbench,
  author       = {Xiaochuan Li and
                  Ryan Ming and
                  Pranav Setlur and
                  Abhijay Paladugu and
                  Andy Tang and
                  Hao Kang and
                  Shuai Shao and
                  Rong Jin and
                  Chenyan Xiong},
  title        = {Benchmark Test-Time Scaling of General {LLM} Agents},
  journal      = {CoRR},
  volume       = {abs/2602.18998},
  year         = {2026},
  url          = {https://doi.org/10.48550/arXiv.2602.18998},
  doi          = {10.48550/ARXIV.2602.18998},
  eprinttype   = {arXiv},
  eprint       = {2602.18998},
  bibsource    = {dblp computer science bibliography, https://dblp.org}
}

@article{balachandran2025inference,
  author       = {Vidhisha Balachandran and
                  Jingya Chen and
                  Lingjiao Chen and
                  Shivam Garg and
                  Neel Joshi and
                  Yash Lara and
                  John Langford and
                  Besmira Nushi and
                  Vibhav Vineet and
                  Yue Wu and
                  Safoora Yousefi},
  title        = {Inference-Time Scaling for Complex Tasks: Where We Stand and What
                  Lies Ahead},
  journal      = {CoRR},
  volume       = {abs/2504.00294},
  year         = {2025},
  url          = {https://doi.org/10.48550/arXiv.2504.00294},
  doi          = {10.48550/ARXIV.2504.00294},
  eprinttype   = {arXiv},
  eprint       = {2504.00294},
  bibsource    = {dblp computer science bibliography, https://dblp.org}
}

@article{erol2026costofpass,
  author       = {Mehmet Hamza Erol and
                  Batu El and
                  Mirac Suzgun and
                  Mert Y{\"{u}}ksekg{\"{o}}n{\"{u}}l and
                  James Zou},
  title        = {Cost-of-Pass: An Economic Framework for Evaluating Language Models},
  journal      = {CoRR},
  volume       = {abs/2504.13359},
  year         = {2025},
  url          = {https://doi.org/10.48550/arXiv.2504.13359},
  doi          = {10.48550/ARXIV.2504.13359},
  eprinttype   = {arXiv},
  eprint       = {2504.13359},
  bibsource    = {dblp computer science bibliography, https://dblp.org}
}

@article{kapoor2025agentsmatter,
  author       = {Sayash Kapoor and
                  Benedikt Stroebl and
                  Zachary S. Siegel and
                  Nitya Nadgir and
                  Arvind Narayanan},
  title        = {{AI} Agents That Matter},
  journal      = {Trans. Mach. Learn. Res.},
  volume       = {2025},
  year         = {2025},
  url          = {https://openreview.net/forum?id=Zy4uFzMviZ},
  bibsource    = {dblp computer science bibliography, https://dblp.org}
}

@article{chen2024frugalgpt,
  author       = {Lingjiao Chen and
                  Matei Zaharia and
                  James Zou},
  title        = {FrugalGPT: How to Use Large Language Models While Reducing Cost and
                  Improving Performance},
  journal      = {Trans. Mach. Learn. Res.},
  volume       = {2024},
  year         = {2024},
  url          = {https://openreview.net/forum?id=cSimKw5p6R},
  bibsource    = {dblp computer science bibliography, https://dblp.org}
}

@inproceedings{ong2025routellm,
  author       = {Isaac Ong and
                  Amjad Almahairi and
                  Vincent Wu and
                  Wei{-}Lin Chiang and
                  Tianhao Wu and
                  Joseph E. Gonzalez and
                  M. Waleed Kadous and
                  Ion Stoica},
  title        = {RouteLLM: Learning to Route LLMs from Preference Data},
  booktitle    = {The Thirteenth International Conference on Learning Representations,
                  {ICLR} 2025, Singapore, April 24-28, 2025},
  publisher    = {OpenReview.net},
  year         = {2025},
  url          = {https://openreview.net/forum?id=8sSqNntaMr},
  bibsource    = {dblp computer science bibliography, https://dblp.org}
}

@article{liang2023helm,
  author       = {Percy Liang and
                  Rishi Bommasani and
                  Tony Lee and
                  Dimitris Tsipras and
                  Dilara Soylu and
                  Michihiro Yasunaga and
                  Yian Zhang and
                  Deepak Narayanan and
                  Yuhuai Wu and
                  Ananya Kumar and
                  Benjamin Newman and
                  Binhang Yuan and
                  Bobby Yan and
                  Ce Zhang and
                  Christian Cosgrove and
                  Christopher D. Manning and
                  Christopher R{\'{e}} and
                  Diana Acosta{-}Navas and
                  Drew A. Hudson and
                  Eric Zelikman and
                  Esin Durmus and
                  Faisal Ladhak and
                  Frieda Rong and
                  Hongyu Ren and
                  Huaxiu Yao and
                  Jue Wang and
                  Keshav Santhanam and
                  Laurel J. Orr and
                  Lucia Zheng and
                  Mert Y{\"{u}}ksekg{\"{o}}n{\"{u}}l and
                  Mirac Suzgun and
                  Nathan Kim and
                  Neel Guha and
                  Niladri S. Chatterji and
                  Omar Khattab and
                  Peter Henderson and
                  Qian Huang and
                  Ryan Chi and
                  Sang Michael Xie and
                  Shibani Santurkar and
                  Surya Ganguli and
                  Tatsunori Hashimoto and
                  Thomas Icard and
                  Tianyi Zhang and
                  Vishrav Chaudhary and
                  William Wang and
                  Xuechen Li and
                  Yifan Mai and
                  Yuhui Zhang and
                  Yuta Koreeda},
  title        = {Holistic Evaluation of Language Models},
  journal      = {Trans. Mach. Learn. Res.},
  volume       = {2023},
  year         = {2023},
  url          = {https://openreview.net/forum?id=iO4LZibEqW},
  bibsource    = {dblp computer science bibliography, https://dblp.org}
}

@inproceedings{sardana2024chinchilla,
  author       = {Nikhil Sardana and
                  Jacob P. Portes and
                  Sasha Doubov and
                  Jonathan Frankle},
  editor       = {Ruslan Salakhutdinov and
                  Zico Kolter and
                  Katherine A. Heller and
                  Adrian Weller and
                  Nuria Oliver and
                  Jonathan Scarlett and
                  Felix Berkenkamp},
  title        = {Beyond Chinchilla-Optimal: Accounting for Inference in Language Model
                  Scaling Laws},
  booktitle    = {Forty-first International Conference on Machine Learning, {ICML} 2024,
                  Vienna, Austria, July 21-27, 2024},
  series       = {Proceedings of Machine Learning Research},
  volume       = {235},
  pages        = {43445--43460},
  publisher    = {{PMLR} / OpenReview.net},
  year         = {2024},
  url          = {https://proceedings.mlr.press/v235/sardana24a.html},
  bibsource    = {dblp computer science bibliography, https://dblp.org}
}

@article{sadhukhan2025kinetics,
  author       = {Ranajoy Sadhukhan and
                  Zhuoming Chen and
                  Haizhong Zheng and
                  Yang Zhou and
                  Emma Strubell and
                  Beidi Chen},
  title        = {Kinetics: Rethinking Test-Time Scaling Laws},
  journal      = {CoRR},
  volume       = {abs/2506.05333},
  year         = {2025},
  url          = {https://doi.org/10.48550/arXiv.2506.05333},
  doi          = {10.48550/ARXIV.2506.05333},
  eprinttype   = {arXiv},
  eprint       = {2506.05333},
  bibsource    = {dblp computer science bibliography, https://dblp.org}
}

@inproceedings{chen2025overthinking,
  author       = {Xingyu Chen and
                  Jiahao Xu and
                  Tian Liang and
                  Zhiwei He and
                  Jianhui Pang and
                  Dian Yu and
                  Linfeng Song and
                  Qiuzhi Liu and
                  Mengfei Zhou and
                  Zhuosheng Zhang and
                  Rui Wang and
                  Zhaopeng Tu and
                  Haitao Mi and
                  Dong Yu},
  editor       = {Aarti Singh and
                  Maryam Fazel and
                  Daniel Hsu and
                  Simon Lacoste{-}Julien and
                  Felix Berkenkamp and
                  Tegan Maharaj and
                  Kiri Wagstaff and
                  Jerry Zhu},
  title        = {Do {NOT} Think That Much for 2+3=? On the Overthinking of Long Reasoning
                  Models},
  booktitle    = {Forty-second International Conference on Machine Learning, {ICML}
                  2025, Vancouver, BC, Canada, July 13-19, 2025},
  series       = {Proceedings of Machine Learning Research},
  volume       = {267},
  publisher    = {{PMLR} / OpenReview.net},
  year         = {2025},
  url          = {https://proceedings.mlr.press/v267/chen25bx.html},
  bibsource    = {dblp computer science bibliography, https://dblp.org}
}

@inproceedings{han2025tale,
  author       = {Tingxu Han and
                  Zhenting Wang and
                  Chunrong Fang and
                  Shiyu Zhao and
                  Shiqing Ma and
                  Zhenyu Chen},
  editor       = {Wanxiang Che and
                  Joyce Nabende and
                  Ekaterina Shutova and
                  Mohammad Taher Pilehvar},
  title        = {Token-Budget-Aware {LLM} Reasoning},
  booktitle    = {Findings of the Association for Computational Linguistics, {ACL} 2025,
                  Vienna, Austria, July 27 - August 1, 2025},
  series       = {Findings of {ACL}},
  volume       = {{ACL} 2025},
  pages        = {24842--24855},
  publisher    = {Association for Computational Linguistics},
  year         = {2025},
  url          = {https://doi.org/10.18653/v1/2025.findings-acl.1274},
  doi          = {10.18653/V1/2025.FINDINGS-ACL.1274},
  bibsource    = {dblp computer science bibliography, https://dblp.org}
}

@article{yue2024cascades,
  author       = {Murong Yue and
                  Jie Zhao and
                  Min Zhang and
                  Liang Du and
                  Ziyu Yao},
  title        = {Large Language Model Cascades with Mixture of Thoughts Representations
                  for Cost-efficient Reasoning},
  journal      = {CoRR},
  volume       = {abs/2310.03094},
  year         = {2023},
  url          = {https://doi.org/10.48550/arXiv.2310.03094},
  doi          = {10.48550/ARXIV.2310.03094},
  eprinttype   = {arXiv},
  eprint       = {2310.03094},
  bibsource    = {dblp computer science bibliography, https://dblp.org}
}

@inproceedings{gupta2024cascades,
  author       = {Neha Gupta and
                  Harikrishna Narasimhan and
                  Wittawat Jitkrittum and
                  Ankit Singh Rawat and
                  Aditya Krishna Menon and
                  Sanjiv Kumar},
  title        = {Language Model Cascades: Token-Level Uncertainty And Beyond},
  booktitle    = {The Twelfth International Conference on Learning Representations,
                  {ICLR} 2024, Vienna, Austria, May 7-11, 2024},
  publisher    = {OpenReview.net},
  year         = {2024},
  url          = {https://openreview.net/forum?id=KgaBScZ4VI},
  bibsource    = {dblp computer science bibliography, https://dblp.org}
}

@inproceedings{aggarwal2024automix,
  author       = {Pranjal Aggarwal and
                  Aman Madaan and
                  Ankit Anand and
                  Srividya Pranavi Potharaju and
                  Swaroop Mishra and
                  Pei Zhou and
                  Aditya Gupta and
                  Dheeraj Rajagopal and
                  Karthik Kappaganthu and
                  Yiming Yang and
                  Shyam Upadhyay and
                  Manaal Faruqui and
                  Mausam},
  editor       = {Amir Globersons and
                  Lester Mackey and
                  Danielle Belgrave and
                  Angela Fan and
                  Ulrich Paquet and
                  Jakub M. Tomczak and
                  Cheng Zhang},
  title        = {AutoMix: Automatically Mixing Language Models},
  booktitle    = {Advances in Neural Information Processing Systems 37: Annual Conference
                  on Neural Information Processing Systems 2024, NeurIPS 2024, Vancouver,
                  BC, Canada, December 10 - 15, 2024},
  year         = {2024},
  url          = {http://papers.nips.cc/paper\_files/paper/2024/hash/ecda225cb187b40ea8edc1f46b03ffda-Abstract-Conference.html},
  bibsource    = {dblp computer science bibliography, https://dblp.org}
}

@inproceedings{ding2024hybrid,
  author       = {Dujian Ding and
                  Ankur Mallick and
                  Chi Wang and
                  Robert Sim and
                  Subhabrata Mukherjee and
                  Victor R{\"{u}}hle and
                  Laks V. S. Lakshmanan and
                  Ahmed Hassan Awadallah},
  title        = {Hybrid {LLM:} Cost-Efficient and Quality-Aware Query Routing},
  booktitle    = {The Twelfth International Conference on Learning Representations,
                  {ICLR} 2024, Vienna, Austria, May 7-11, 2024},
  publisher    = {OpenReview.net},
  year         = {2024},
  url          = {https://openreview.net/forum?id=02f3mUtqnM},
  bibsource    = {dblp computer science bibliography, https://dblp.org}
}

@inproceedings{dekoninck2025cascade,
  author       = {Jasper Dekoninck and
                  Maximilian Baader and
                  Martin T. Vechev},
  editor       = {Aarti Singh and
                  Maryam Fazel and
                  Daniel Hsu and
                  Simon Lacoste{-}Julien and
                  Felix Berkenkamp and
                  Tegan Maharaj and
                  Kiri Wagstaff and
                  Jerry Zhu},
  title        = {A Unified Approach to Routing and Cascading for LLMs},
  booktitle    = {Forty-second International Conference on Machine Learning, {ICML}
                  2025, Vancouver, BC, Canada, July 13-19, 2025},
  series       = {Proceedings of Machine Learning Research},
  volume       = {267},
  publisher    = {{PMLR} / OpenReview.net},
  year         = {2025},
  url          = {https://proceedings.mlr.press/v267/dekoninck25a.html},
  bibsource    = {dblp computer science bibliography, https://dblp.org}
}

@inproceedings{ding2025bestroute,
  author       = {Dujian Ding and
                  Ankur Mallick and
                  Shaokun Zhang and
                  Chi Wang and
                  Daniel Madrigal and
                  Mirian del Carmen Hipolito Garcia and
                  Menglin Xia and
                  Laks V. S. Lakshmanan and
                  Qingyun Wu and
                  Victor R{\"{u}}hle},
  editor       = {Aarti Singh and
                  Maryam Fazel and
                  Daniel Hsu and
                  Simon Lacoste{-}Julien and
                  Felix Berkenkamp and
                  Tegan Maharaj and
                  Kiri Wagstaff and
                  Jerry Zhu},
  title        = {BEST-Route: Adaptive {LLM} Routing with Test-Time Optimal Compute},
  booktitle    = {Forty-second International Conference on Machine Learning, {ICML}
                  2025, Vancouver, BC, Canada, July 13-19, 2025},
  series       = {Proceedings of Machine Learning Research},
  volume       = {267},
  publisher    = {{PMLR} / OpenReview.net},
  year         = {2025},
  url          = {https://proceedings.mlr.press/v267/ding25d.html},
  bibsource    = {dblp computer science bibliography, https://dblp.org}
}

@inproceedings{aggarwal2023adaptive,
  author       = {Pranjal Aggarwal and
                  Aman Madaan and
                  Yiming Yang and
                  Mausam},
  editor       = {Houda Bouamor and
                  Juan Pino and
                  Kalika Bali},
  title        = {Let's Sample Step by Step: Adaptive-Consistency for Efficient Reasoning
                  and Coding with LLMs},
  booktitle    = {Proceedings of the 2023 Conference on Empirical Methods in Natural
                  Language Processing, {EMNLP} 2023, Singapore, December 6-10, 2023},
  pages        = {12375--12396},
  publisher    = {Association for Computational Linguistics},
  year         = {2023},
  url          = {https://doi.org/10.18653/v1/2023.emnlp-main.761},
  doi          = {10.18653/V1/2023.EMNLP-MAIN.761},
  bibsource    = {dblp computer science bibliography, https://dblp.org}
}

@inproceedings{li2024esc,
  author       = {Yiwei Li and
                  Peiwen Yuan and
                  Shaoxiong Feng and
                  Boyuan Pan and
                  Xinglin Wang and
                  Bin Sun and
                  Heda Wang and
                  Kan Li},
  title        = {Escape Sky-high Cost: Early-stopping Self-Consistency for Multi-step
                  Reasoning},
  booktitle    = {The Twelfth International Conference on Learning Representations,
                  {ICLR} 2024, Vienna, Austria, May 7-11, 2024},
  publisher    = {OpenReview.net},
  year         = {2024},
  url          = {https://openreview.net/forum?id=ndR8Ytrzhh},
  bibsource    = {dblp computer science bibliography, https://dblp.org}
}

@inproceedings{trivedi2024appworld,
  author       = {Harsh Trivedi and
                  Tushar Khot and
                  Mareike Hartmann and
                  Ruskin Manku and
                  Vinty Dong and
                  Edward Li and
                  Shashank Gupta and
                  Ashish Sabharwal and
                  Niranjan Balasubramanian},
  editor       = {Lun{-}Wei Ku and
                  Andre Martins and
                  Vivek Srikumar},
  title        = {AppWorld: {A} Controllable World of Apps and People for Benchmarking
                  Interactive Coding Agents},
  booktitle    = {Proceedings of the 62nd Annual Meeting of the Association for Computational
                  Linguistics (Volume 1: Long Papers), {ACL} 2024, Bangkok, Thailand,
                  August 11-16, 2024},
  pages        = {16022--16076},
  publisher    = {Association for Computational Linguistics},
  year         = {2024},
  url          = {https://doi.org/10.18653/v1/2024.acl-long.850},
  doi          = {10.18653/V1/2024.ACL-LONG.850},
  bibsource    = {dblp computer science bibliography, https://dblp.org}
}

@article{tbench2026,
  author       = {Mike A. Merrill and
                  Alexander Glenn Shaw and
                  Nicholas Carlini and
                  Boxuan Li and
                  Harsh Raj and
                  Ivan Bercovich and
                  Lin Shi and
                  Jeong Yeon Shin and
                  Thomas Walshe and
                  Estefany Kelly Buchanan and
                  Junhong Shen and
                  Guanghao Ye and
                  Haowei Lin and
                  Jason Poulos and
                  Maoyu Wang and
                  Marianna Nezhurina and
                  Jenia Jitsev and
                  Di Lu and
                  Orfeas Menis{-}Mastromichalakis and
                  Zhiwei Xu and
                  Zizhao Chen and
                  Yue Liu and
                  Robert Zhang and
                  Leon Liangyu Chen and
                  Anurag Kashyap and
                  Jan{-}Lucas Uslu and
                  Jeffrey Li and
                  Jianbo Wu and
                  Minghao Yan and
                  Song Bian and
                  Vedang Sharma and
                  Ke Sun and
                  Steven Dillmann and
                  Akshay Anand and
                  Andrew Lanpouthakoun and
                  Bardia Koopah and
                  Changran Hu and
                  Etash Kumar Guha and
                  Gabriel H. S. Dreiman and
                  Jiacheng Zhu and
                  Karl Krauth and
                  Li Zhong and
                  Niklas Muennighoff and
                  Robert Amanfu and
                  Shangyin Tan and
                  Shreyas Pimpalgaonkar and
                  Tushar Aggarwal and
                  Xiangning Lin and
                  Xin Lan and
                  Xuandong Zhao and
                  Yiqing Liang and
                  Yuanli Wang and
                  Zilong Wang and
                  Changzhi Zhou and
                  David Heineman and
                  Hange Liu and
                  Harsh Trivedi and
                  John Yang and
                  Junhong Lin and
                  Manish Shetty and
                  Michael Yang and
                  Nabil Omi and
                  Negin Raoof and
                  Shanda Li and
                  Terry Yue Zhuo and
                  Wuwei Lin and
                  Yiwei Dai and
                  Yuxin Wang and
                  Wenhao Chai and
                  Shang Zhou and
                  Dariush Wahdany and
                  Ziyu She and
                  Jiaming Hu and
                  Zhikang Dong and
                  Yuxuan Zhu and
                  Sasha Cui and
                  Ahson Saiyed and
                  Arinbj{\"{o}}rn Kolbeinsson and
                  Jesse Hu and
                  Christopher Michael Rytting and
                  Ryan Marten and
                  Yixin Wang and
                  Alex Dimakis and
                  Andy Konwinski and
                  Ludwig Schmidt},
  title        = {Terminal-Bench: Benchmarking Agents on Hard, Realistic Tasks in Command
                  Line Interfaces},
  journal      = {CoRR},
  volume       = {abs/2601.11868},
  year         = {2026},
  url          = {https://doi.org/10.48550/arXiv.2601.11868},
  doi          = {10.48550/ARXIV.2601.11868},
  eprinttype   = {arXiv},
  eprint       = {2601.11868},
  bibsource    = {dblp computer science bibliography, https://dblp.org}
}

@misc{harbor2026,
author = {{Harbor Framework Team}},
title = {{Harbor: A framework for evaluating and optimizing agents and models in container environments}},
year = {2026},
doi = {10.5281/zenodo.20953922},
url = {https://doi.org/10.5281/zenodo.20953922}
}

@inproceedings{jain2025livecodebench,
  author       = {Naman Jain and
                  King Han and
                  Alex Gu and
                  Wen{-}Ding Li and
                  Fanjia Yan and
                  Tianjun Zhang and
                  Sida Wang and
                  Armando Solar{-}Lezama and
                  Koushik Sen and
                  Ion Stoica},
  title        = {LiveCodeBench: Holistic and Contamination Free Evaluation of Large
                  Language Models for Code},
  booktitle    = {The Thirteenth International Conference on Learning Representations,
                  {ICLR} 2025, Singapore, April 24-28, 2025},
  publisher    = {OpenReview.net},
  year         = {2025},
  url          = {https://openreview.net/forum?id=chfJJYC3iL},
  bibsource    = {dblp computer science bibliography, https://dblp.org}
}

@inproceedings{gao2025omnimath,
  author       = {Bofei Gao and
                  Feifan Song and
                  Zhe Yang and
                  Zefan Cai and
                  Yibo Miao and
                  Qingxiu Dong and
                  Lei Li and
                  Chenghao Ma and
                  Liang Chen and
                  Runxin Xu and
                  Zhengyang Tang and
                  Benyou Wang and
                  Daoguang Zan and
                  Shanghaoran Quan and
                  Ge Zhang and
                  Lei Sha and
                  Yichang Zhang and
                  Xuancheng Ren and
                  Tianyu Liu and
                  Baobao Chang},
  title        = {Omni-MATH: {A} Universal Olympiad Level Mathematic Benchmark for Large
                  Language Models},
  booktitle    = {The Thirteenth International Conference on Learning Representations,
                  {ICLR} 2025, Singapore, April 24-28, 2025},
  publisher    = {OpenReview.net},
  year         = {2025},
  url          = {https://openreview.net/forum?id=yaqPf0KAlN},
  bibsource    = {dblp computer science bibliography, https://dblp.org}
}

@inproceedings{balunovic2025matharena,
  author       = {Mislav Balunovic and
                  Jasper Dekoninck and
                  Ivo Petrov and
                  Nikola Jovanovic and
                  Martin T. Vechev},
  editor       = {Danielle Belgrave and
                  Cheng Zhang and
                  Laura N. Montoya and
                  Hsuan{-}Tien Lin and
                  Razvan Pascanu and
                  Piotr Koniusz and
                  Marzyeh Ghassemi and
                  Nancy Chen and
                  Iv{\'{a}}n Vladimir Meza Ru{\'{\i}}z and
                  Arturo Loaiza{-}Bonilla},
  title        = {MathArena: Evaluating LLMs on Uncontaminated Math Competitions},
  booktitle    = {Advances in Neural Information Processing Systems 38: Annual Conference
                  on Neural Information Processing Systems 2025, NeurIPS 2025, San Diego,
                  CA, USA, December 2-7, 2025 / Mexico City, Mexico, November 30 - December
                  5, 2025},
  year         = {2025},
  url          = {http://papers.nips.cc/paper\_files/paper/2025/hash/1d27c01ebd3e3aebe226b44fc970d803-Abstract-Datasets\_and\_Benchmarks\_Track.html},
  bibsource    = {dblp computer science bibliography, https://dblp.org}
}

@misc{matharena2025apex,
  title        = {{MathArena Apex}},
  author       = {{MathArena}},
  howpublished = {\url{https://matharena.ai/apex/}},
  year         = {2025}
}

@inproceedings{liu2025compassverifier,
  author       = {Shudong Liu and
                  Hongwei Liu and
                  Junnan Liu and
                  Linchen Xiao and
                  Songyang Gao and
                  Chengqi Lyu and
                  Yuzhe Gu and
                  Wenwei Zhang and
                  Derek F. Wong and
                  Songyang Zhang and
                  Kai Chen},
  editor       = {Christos Christodoulopoulos and
                  Tanmoy Chakraborty and
                  Carolyn Rose and
                  Violet Peng},
  title        = {CompassVerifier: {A} Unified and Robust Verifier for LLMs Evaluation
                  and Outcome Reward},
  booktitle    = {Proceedings of the 2025 Conference on Empirical Methods in Natural
                  Language Processing, {EMNLP} 2025, Suzhou, China, November 4-9, 2025},
  pages        = {33466--33494},
  publisher    = {Association for Computational Linguistics},
  year         = {2025},
  url          = {https://doi.org/10.18653/v1/2025.emnlp-main.1698},
  doi          = {10.18653/V1/2025.EMNLP-MAIN.1698},
  bibsource    = {dblp computer science bibliography, https://dblp.org}
}

\clearpage
\appendix
\etocdepthtag.toc{appendix}
\etocsettagdepth{mainmatter}{none}
\etocsettagdepth{appendix}{subsection}
\etocsettocstyle{%
  \begin{center}\sffamily\bfseries\fontsize{16}{19}\selectfont Appendix\end{center}
  \vskip 2mm {\color{rulegray}\hrule height 0.4pt}\vskip 4mm\setlength{\parskip}{3pt}}{%
  \vskip 3mm {\color{rulegray}\hrule height 0.4pt}}
\tableofcontents
\clearpage

\section{Budgets and caps}
\label{app:budgets}

\emph{Each domain has one unit budget $B$ and a set of hard caps; a protocol at multiplier $m$ receives $m$ times every cap.}

\begin{table}[h]
\centering\small
\caption{Nominal stage allowances for static tasks. The suffix denotes a total budget multiplier, not a sample count. Provider limits can reduce individual allowances; in particular, Qwen3-Next's revision is capped at $B$, giving at most $3B$ across its three stages.}
\label{tab:stages}
\begin{tabular*}{\linewidth}{@{\extracolsep{\fill}}lllc@{}}
\toprule
Protocol & Generation & Selection or feedback & Total cap \\
\midrule
Direct & $B$ & -- & $B$ \\
Parallel@2 & $2\times B$ & rule, no model tokens & $2B$ \\
Parallel@4 & $3\times B$ & joint selection, $B$ & $4B$ \\
Revise@4 & initial $B$, revision up to $2B$ & critique, $B$ & $4B$ \\
\bottomrule
\end{tabular*}
\end{table}

Table~\ref{tab:stages} and Algorithm~\ref{alg:static} specify the static allocations.
Parallel@4 and Revise@4 share an upper budget envelope, but their accuracy difference does not isolate a mechanism at identical realized cost.

\begin{table}[h]
\centering
\small
\caption{Per-cell caps for Direct ($m=1$). All caps are multiplied by $m$ for Parallel@2 ($m=2$), Parallel@4 and Revise@4 ($m=4$).}
\label{tab:app-budgets}
\begin{tabular*}{\linewidth}{@{\extracolsep{\fill}}lrrrr@{}}
\toprule
Domain & Output tokens $B$ & Model calls & Tool calls & Wall clock \\
\midrule
Math & 131{,}072 & 16 & -- & 2 h \\
Code & 131{,}072 & 64 & 128 & 2 h \\
Workflow & 393{,}216\textsuperscript{a} & 48 & 64 & 1 h \\
Terminal & 393{,}216 & 128 & 256 & 8 h\textsuperscript{b} \\
\bottomrule
\end{tabular*}

\smallskip
\raggedright\footnotesize
\textsuperscript{a}At most 131{,}072 per executor turn.
\textsuperscript{b}The native task limit; a single shell command is additionally limited to 1{,}800\,s (Appendix~\ref{app:harness}).
\end{table}

Within agent protocols, each plan may use up to 32{,}768 output tokens, plan selection up to 65{,}536, and each review up to 131{,}072, and a review call is never capped below an executor turn; all stages share the cell's total allowance.
The mathematics allowance exceeds the longest Flash Direct solution observed during development (94{,}834 tokens); one of Flash's 60 mathematics Direct cells on the test set was truncated. This headroom does not guarantee that every provider-specific stage is untruncated.

\clearpage  
\section{Test set construction}
\label{app:testset}

\emph{Tasks were admitted by a model-independent eligibility check, drawn by stratum with a fixed seed, and sealed before any test-time call.}

\paragraph{Eligibility.}
Where an executable reference is distributed, eligibility requires that it succeeds and a blank or no-op submission fails under the pinned runtime. AppWorld uses the separate contract below because its test references are unavailable.
For code, an official editorial program must pass every hidden test in our container; duplicate references are removed by code hash.
AppWorld's test split ships no reference solutions, so eligibility there requires the official development examples to pass, the native evaluator chain to be fixed, and a blank state to fail; all 585 test tasks met this contract.
For Terminal-Bench, the Harbor \texttt{oracle} agent must obtain reward 1 and the \texttt{nop} agent reward 0; 62 of 66 tasks qualified, three were excluded for requiring GPUs, and one because its reference could not run in the shipped image.

\paragraph{Sampling and exclusions.}
Within each source, tasks were stratified by published difficulty and sampled without replacement using seed 20260917, taking at most one task per scenario cluster or leakage group.
Every task seen by any model during development (math 52, code 25, workflow 15, terminal 1) was excluded, together with three AtCoder contests held out as a whole (abc391, abc398, arc191).
The set was sealed on 2026-09-21; the seal records hashes of the specification and of the public and private task files, and the tasks did not change afterwards.
Table~\ref{tab:domains} in Section~\ref{sec:method-domains} summarizes the four domains, and Table~\ref{tab:app-strata} gives their composition by stratum.

\begin{table}[h]
\centering
\small
\caption{Composition of the sealed test set.}
\label{tab:app-strata}
\resizebox{\linewidth}{!}{%
\begin{tabular}{@{}lll@{}}
\toprule
Domain & Source & Strata (tasks) \\
\midrule
Math (60) & APEX shortlist; APEX 2025 (MathArena) & 27; 7 \\
 & Omni-MATH-Rule, difficulty $[3.5,5)$ / $[5,6)$ / $[6,7)$ / $[7,10)$ & 8 / 8 / 5 / 5 \\
Code (100) & LiveCodeBench v6, AtCoder, 2023-08 to 2025-04 & easy / medium / hard: 30 / 28 / 42 \\
Workflow (160) & AppWorld test (data 0.1.0), normal, levels 1 / 2 / 3 & 17 / 14 / 19 \\
 & AppWorld test, challenge, levels 1 / 2 / 3 & 19 / 40 / 51 \\
Terminal (30) & Terminal-Bench 4.0.0 via Harbor & easy / medium / hard: 7 / 11 / 12 \\
\bottomrule
\end{tabular}}
\end{table}

The programming test set contains 42 hard, 28 medium, and 30 easy tasks. Its composition was fixed before test inference; development performance informed the source and quota choices, not the selection of individual held-out items by observed TTS gain.

\clearpage  
\section{Protocol pseudocode}
\label{app:algorithms}

\begin{algorithm}[h]
\caption{Static protocols. $\mathcal{M}(\cdot\,;b)$ denotes a generation stage with output allowance $b$; on code it can include multiple calls with public-tool feedback. Provider limits can reduce a stage allowance. Identical requests may reuse stored samples across protocols, charged logically to each.}
\label{alg:static}
\begin{algorithmic}[1]
\Function{Direct}{$x$}
  \State \Return $\mathcal{M}(x;\,B)$
\EndFunction
\Function{Parallel@2}{$x$}
  \State $y_1, y_2 \gets \mathcal{M}(x;\,B)$ \Comment{independent draws}
  \State \Return $\textsc{Rule}(y_1, y_2)$ \Comment{no model tokens}
\EndFunction
\Function{Parallel@4}{$x$}
  \State $y_1, y_2, y_3 \gets \mathcal{M}(x;\,B)$
  \State $j \gets \mathcal{M}(\textsc{Judge}(x, y_1, y_2, y_3);\,B)$ \Comment{same model reads all three}
  \State \Return $y_j$
\EndFunction
\Function{Revise@4}{$x$}
  \State $y_0 \gets \mathcal{M}(x;\,B)$
  \State $f \gets \mathcal{M}(\textsc{Critique}(x, y_0);\,B)$ \Comment{feedback only, no new answer}
  \If{$f = \texttt{PASS}$} \State \Return $y_0$ \EndIf
  \State \Return $\mathcal{M}(\textsc{Revise}(x, y_0, f);\,2B)$
\EndFunction
\end{algorithmic}
\end{algorithm}

\begin{algorithm}[h]
\caption{Protocols on agent tasks. $o_0$ is the initial observation; the whole episode obeys Eq.~\eqref{eq:budget}.}
\label{alg:agent}
\begin{algorithmic}[1]
\Function{Parallel@$k$}{$x$}
  \State $p_i \gets \mathcal{M}(\textsc{Plan}(x, o_0))$ for $i = 1,\dots,k$ \Comment{no tool calls}
  \State $j \gets \mathcal{M}(\textsc{Choose}(p_1,\dots,p_k))$
  \State \Return $\textsc{Execute}(x, p_j)$ \Comment{a single episode}
\EndFunction
\Function{Revise@4}{$x$}
  \State $h \gets (x, o_0)$
  \While{the executor issues a tool call and the budget allows}
    \If{two turns have passed since the last review}
      \State $h \gets h \,\Vert\, \mathcal{M}(\textsc{Review}(h))$ \Comment{continue, repair, replan, or stop}
    \EndIf
    \State $h \gets h \,\Vert\, \textsc{Step}(\mathcal{M}, h)$ \Comment{one tool call and its observation}
  \EndWhile
  \State \Return final environment state
\EndFunction
\end{algorithmic}
\end{algorithm}

\clearpage  
\section{Protocol details}
\label{app:protocols}

\emph{The original protocols require a formatted choice and use no fallback after an unparseable selection. Appendix~\ref{app:selector-replay} evaluates output-preserving alternatives on the same candidates.}

\paragraph{Joint selection (Parallel@4, static tasks).}
The model receives the problem and the three candidates, labelled A to C, and is told to verify each candidate's decisive steps rather than compare their popularity.
It must end with \texttt{SELECTED: <letter>}.
If that line is missing but the last boxed expression is a single candidate letter, the choice is accepted and the parse route is recorded, so the strict score remains recomputable; any other reply is a format failure.

\paragraph{Self-Refine (Revise@4, static tasks).}
The critique prompt forbids a replacement solution and a final answer, asks the model to quote each faulty step and state what should change, and requires \texttt{VERDICT: PASS} or \texttt{VERDICT: REVISE}.
A \texttt{PASS} returns the initial solution without a revision call; an unparseable verdict proceeds to revision.
If critique fails to finish or revision yields no completed final answer, the frozen harness returns an empty submission without falling back to the initial answer. Appendix~\ref{app:revision-fallback} evaluates this implementation choice offline.
Whether the feedback asserts an answer different from the initial one is recorded per cell.
Relative to \citet{madaan2023selfrefine}, we run one feedback--refine iteration and no few-shot demonstrations.

\paragraph{Public-test selection (Parallel@2, code).}
Each candidate program is compiled and run on the public examples; the one matching the most examples is kept, with ties going to the earlier candidate. Compile failures or empty programs receive a penalty below a runnable program that matches no public examples.
Hidden tests are never consulted, and the step consumes no model tokens, although it requires executing the candidates.

\paragraph{Agent protocols.}
For Parallel@$k$, the model proposes $k$ plans from the initial observation without calling tools, then replies \texttt{SELECT: N}; an invalid selection ends the cell as a stage failure.
The chosen plan is appended to the context and executed once.
For Revise@4, after every second executor turn the model is asked to inspect the goal, the visible state, and recent errors, to choose among continuing, inspecting and repairing, replanning, or stopping, to preserve actions that already succeeded, and not to claim success without checking the observable state.
The review text is appended to the context; the program does not parse the chosen option, and the episode ends when the executor issues no further tool call.

\paragraph{Protocols not included.}
A separately instructed longer-reasoning protocol was not run. Direct already uses high effort and a generous allowance, but this does not establish equivalence to all alternative long-reasoning prompts or provider effort settings.
Revise at $2\times$ is excluded by our choice to allocate a full solution allowance to each of its three stages; other revision allocations remain possible.
Domain-specific methods, such as moderated debate for mathematics or test-driven repair for code, were planned as a separate audit and are not part of the reported results.

\clearpage  
\section{Grading and runtime}
\label{app:grading}

\emph{Static outputs use the declared answer or execution grader, and agents use native final-state evaluators; hidden answers and tests are used only for offline grading.}

\begin{itemize}
  \item \textbf{Math.} A rule-based expression grader checks the final scalar answer. An offline CompassVerifier-3B model~\citep{liu2025compassverifier} grades the same outputs as a second opinion; the two disagree on 25 of 1{,}069 scored mathematics cells.
  \item \textbf{Code.} Programs run against all hidden tests in a pinned container (image \texttt{sha256:4a33fba9\ldots}; CPython 3.11; g++ 12.2 with \texttt{-std=c++20}; AtCoder Library). Re-running eligibility for all 282 AtCoder candidates under C++20 in the same image added six tasks and removed none.
  \item \textbf{Workflow.} AppWorld's native evaluator reports task goal completion, which requires every check to pass; we do not call it the official scenario goal completion.
  \item \textbf{Terminal.} Terminal-Bench's native verifier, run by Harbor in a separate container, returns a binary reward.
\end{itemize}

Following AppWorld's usage terms, we report only aggregate results on its test split and do not inspect individual test trajectories.

\clearpage  
\section{Agent harness}
\label{app:harness}

\emph{Agents act through one uniform tool interface per domain, and the harness enforces limits inside the task environment.}

Each cell begins from the prescribed initial state, with no memory carried across tasks.
Agents can change the environment and may sometimes undo individual actions, but the harness forbids branching into cloned worlds, resetting an episode, or selecting among multiple executed worlds.

Terminal tasks run in their native Harbor~\citep{harbor2026} containers through a single shell-command tool rather than the tmux interface of the reference Terminus-2 agent, so our scores are not comparable to the public leaderboard.
A single command is limited to 1{,}800\,s, the longest native verifier timeout in our terminal set.
The limit is enforced inside the container and terminates the whole process group; a stopped command returns exit status 124 to the model, and the tool description states the limit.

The AppWorld environment server serves one episode at a time.
Half of the 160 workflow runs used four replica servers that match the original in image, read-only data, and resource limits and differ only in port; each run records the server it used.

\clearpage  
\section{Development evidence behind the design}
\label{app:dev-evidence}

\emph{Three design rules---full budgets for judges, a judge at the solver's effort, and verification instead of voting in Parallel@4---were adopted after development measurements, before the test set was sealed.}

\begin{itemize}
  \item \textbf{Observed judging costs.} A critique at low effort used 58{,}173 output tokens, 58{,}100 of them reasoning, to emit 255 characters; critiques capped at $B/8$ or $B/16$ stopped mid-reasoning and returned no verdict. Selection calls used 14{,}637--61{,}719 output tokens.
  \item \textbf{A low-effort critique error.} On a problem the solver answered correctly in three of three attempts, a low-effort critique asserted a different answer as established.
  \item \textbf{Development selector comparison.} On 16 development problems, Direct scored 0.625, majority voting over three samples 0.562, and at least one of three samples was correct 81\% of the time. On the seven contested problems, joint verification recovered four of five that had a correct candidate; voting recovered one. Self-verification accepted a wrong solution in 7 of 15 usable judgments.
  \item \textbf{Eight-sample pilot.} On the hardest seven development problems with eight samples each, voting peaked at four samples while coverage kept rising.
\end{itemize}
These small development observations motivated the original design; they do not establish that judging is preferable on held-out tasks. The same-pool test analysis in Appendix~\ref{app:selector-replay} measures that marginal value explicitly.

\clearpage  
\section{Accounting, failures, and changes after sealing}
\label{app:accounting}

\emph{We distinguish logical method cost from campaign spending, task correctness from termination status, and a fixed task seal from versioned implementation amendments.}

\paragraph{Cost ledger.}
Before each call, the ledger reserves the stage cap at the configured price and settles actual usage afterwards; a failed call keeps its reservation. This is conservative for retained calls, but configuration means are not an upper bound on whole-campaign expense: they average retained final cell records and omit superseded infrastructure attempts and records that never completed. Identical requests reused across protocols are charged logically in every configuration, although they are physically paid for only once.
DeepSeek V4.1 Flash uses one fixed peak-rate snapshot for every call: \$0.30 per million uncached input tokens, \$1.20 per million output tokens, and \$0.006 per million cache-read tokens. Token counts and cache hits come from the recorded usage; prices do not depend on the call timestamp. The half-price sensitivity multiplies every DeepSeek configuration cost by 0.5, holding usage fixed. Neither scenario reconstructs the actual invoice, although most calls ran off-peak.
Bedrock models are charged at list price; Opus~5.5 agent requests use prompt caching, which changes cost but not model inputs or outputs.
The GPT-OSS price on the pricing page is the Sydney region's, so its ledger errs high.
An accounting audit recovered 12 Flash terminal cell records whose parent runs had not written an aggregate evaluation file. Their grades were already retained, but the merger had defaulted their costs to zero. We restore the saved costs (\$6.20 in total) and token usage, leaving all outcomes unchanged: Direct, Parallel@2, and Parallel@4 cost \$0.307, \$0.363, and \$0.438 per retained task, respectively. The tables and Figure~\ref{fig:pareto} use these corrected values; \texttt{analysis/audit\_terminal\_costs.py} records the source hashes and correction. Terminal frontier membership is unchanged.
Ledger totals are about \$101 for DeepSeek (including screening), \$282 for Qwen3-Next and GLM-5, \$83 for Opus~5.5, and \$82 for GPT-OSS and MiniMax.

\paragraph{Failures.}
Cells with no usable submission or no final record score zero, and model outcomes are not retried to obtain a better answer. For agents, the native evaluator scores the final world state independently of the termination status. The retained matrix includes six correct resource-limited cells: two Flash workflow Direct cells, one Flash terminal Direct cell, one Flash terminal Parallel@4 cell, and two Opus terminal Direct cells. Reporting every non-completed status as incorrect would therefore misdescribe the primary scores.
Only infrastructure failures---dropped connections, provider server errors, permission errors, and task containers that failed to start---were resumed or re-run, at most twice per task.
Fifteen cells never finished and are scored as incorrect: two Flash terminal cells that reached their wall-clock limit, ten GLM-5 mathematics cells whose Bedrock requests hung until the client timeout (about 5\% of GLM-5 requests hung; every successful call finished within 1{,}798\,s), and three GPT-OSS cells whose identical request hung or returned a server error on every retry.
Six terminal Direct cells whose Harbor container failed to start or whose command exceeded the operating system's argument limit were re-run and now have valid grades; two verifier timeouts count as failures, as Terminal-Bench scores them.

\paragraph{Opus~5.5 on code without numpy.}
Six of Opus~5.5's seven code failures import numpy, which the frozen grading image does not contain, so these programs fail at import.
The prompt states no library restriction, but it offers a tool that runs code in the same sandbox, and Opus~5.5 submitted these six programs without running them; no other model imported numpy.
We therefore keep the official score of 0.930.
Regrading the six programs in the same image with numpy~2.4.6 installed and \texttt{OPENBLAS\_NUM\_THREADS=1} set makes all six pass, giving 0.990.

\paragraph{Changes after sealing.}
Four infrastructure changes were made after sealing, and each affected configuration declares it.
(i) Terminal tasks run through a native Harbor bridge, including the per-command limit.
(ii) Amazon Bedrock's Converse API rejected three agent-loop situations as validation errors, ending runs for interface reasons: a tool-less review call over a transcript with tool turns, a call to a tool name the model invented, and tool calls written as plain JSON text. The first two are now rendered as text, and calls to declared tools written as text are recovered and marked (24\% of Qwen3-Next's workflow calls and 11\% of its terminal calls).
(iii) Prompt caching on Bedrock agent requests.
(iv) Output caps of selection stages fitted to the remaining context window for models whose window is smaller than the frozen cap.
These changes preserve the sealed tasks and primary graders, but they affect tool-history serialization, executable tool handling, or the effective per-request cap. Nominal cell allowances do not imply identical realized inputs and stage limits across interface versions. The Qwen3-Next runs on the earlier interface are retained as an implementation baseline, not pooled into the final matrix.

\clearpage  
\section{Models}
\label{app:models}

\emph{Solver and review stages request high reasoning effort; agent planning and plan selection request low effort. Provider-specific limits and interface settings are part of the reported configuration.}

Table~\ref{tab:models} in Section~\ref{sec:method-models} lists access, prices, and context windows.
Only Flash and Qwen3-Next meet the context-headroom requirement of our full-solution critique configuration; this exclusion does not imply that shorter or differently structured revision is impossible for the other models.
DeepSeek V4.1 Flash is accessed through the official API with thinking enabled, reasoning effort high, and $\text{top-}p = 0.95$; no temperature or seed is sent.
The other models are accessed through Amazon Bedrock (us-east-1; Opus~5.5 through the global endpoint with adaptive thinking at effort high).
We accept a single-call output limit only when a generation actually stops at that limit, because provider model cards understated capacity for four of eight candidates during screening.
Bedrock caps output per call at 128{,}000 tokens for Opus~5.5 and GPT-OSS, below the nominal 131{,}072; affected stages are capped explicitly while total cell allowances remain unchanged. Qwen3-Next's static revision is capped at 131{,}072 rather than the nominal 262{,}144, so its three-stage allocation totals at most $3B$ under a $4B$ cell allowance.
GPT-OSS's 131{,}072-token window additionally limits its solution stages to 118{,}000 tokens, which truncated three mathematics candidates, and its selection stages are fitted to the remaining window.
On one terminal task, Bedrock's content filter stopped two of Opus~5.5's turns, ending the episode; it counts as a model outcome.

\clearpage  
\section{Offline diagnostic and statistical audit}
\label{app:offline-audit}
\suppressfloats[t]

The audit in \texttt{analysis/audit\_results.py} validates the number of retained records and correct outcomes for all 68 configurations against the frozen result matrix before computing diagnostics.
It also checks that each task index agrees with the sealed public manifest.
Only retained final runs and declared parts are merged; superseded attempts are excluded.
AppWorld records are projected to aggregate status and grade counts without inspecting individual trajectories.
The output contains aggregate statistics and source hashes, not task text or model answers.

\paragraph{Dependence and uncertainty.}
For each domain--protocol panel, the bootstrap resamples task identifiers within their frozen strata, taking all evaluated models' paired outcomes for each sampled task together.
Models are fixed rather than sampled from a population.
We use 20{,}000 replicates with seed 20260925 and percentile 95\% intervals.
Per-model exact sign tests are separately corrected across all 44 evaluated non-Direct comparisons using Holm's procedure (Table~\ref{tab:holm}); the twelve panel intervals are pointwise descriptive intervals and are not multiplicity-adjusted significance tests.
The absence of repeated seeds prevents estimating within-task stochastic variability.

\begin{table}[t]
\centering\small
\caption{Per-model comparisons with Holm-adjusted $p<0.05$ across the full family of 44 protocol-versus-Direct tests. The family includes every evaluated non-Direct configuration, not only displayed effects.}
\label{tab:holm}
\begin{tabular}{@{}lllrr@{}}
\toprule
Model & Domain & Protocol & Wins--losses & Adjusted $p$ \\
\midrule
MiniMax M2.5 & Workflow & Parallel@4 & 0--19 & 0.0001678 \\
MiniMax M2.5 & Workflow & Parallel@2 & 1--19 & 0.001722 \\
GPT-OSS-120B & Code & Parallel@4 & 4--26 & 0.002498 \\
GLM-5 & Code & Parallel@2 & 15--0 & 0.002502 \\
DeepSeek V4.1 Flash & Workflow & Revise@4 & 11--34 & 0.03296 \\
\bottomrule
\end{tabular}
\end{table}

\paragraph{Reused outputs and revision transitions.}
Request caching couples some static protocol outputs, but tool continuations or split runs can produce different initial strings.
For example, Flash's coding revision initial output is byte-identical to Direct on 88/100 tasks, although its initial correctness agrees on all 100.
Revision transitions are consequently computed from each cell's own saved initial and final output, not inferred from cross-protocol labels.
The four static transition rows in Table~\ref{tab:revisiontransitions} reproduce the reported revision accuracy changes.
The same interpretation does not extend to separately executed agent episodes.

\begin{table}[t]
\centering\small
\caption{Static Revise@4 transitions from its own recorded initial answer to its final submission. R and W denote correct and incorrect, including missing or ungradable outputs. Recovery and loss are operational rates, not solely content changes.}
\label{tab:revisiontransitions}
\begin{tabular}{@{}llrrrrrr@{}}
\toprule
Model & Domain & R$\to$R & W$\to$R & R$\to$W & W$\to$W & Recovery & Loss \\
\midrule
DeepSeek V4.1 Flash & Math & 46 & 2 & 0 & 12 & 2/14 & 0/46 \\
Qwen3-Next-80B & Math & 30 & 0 & 1 & 29 & 0/29 & 1/31 \\
DeepSeek V4.1 Flash & Code & 98 & 0 & 1 & 1 & 0/1 & 1/99 \\
Qwen3-Next-80B & Code & 76 & 0 & 0 & 24 & 0/24 & 0/76 \\
\bottomrule
\end{tabular}
\smallskip
\parbox{\linewidth}{\footnotesize Flash mathematics has one repair and one recovery from an empty, truncated initial output. Flash code loses a correct initial program after critique truncation, without a revised submission. Qwen mathematics has one correct-to-incorrect rewrite. See Appendix~\ref{app:revision-fallback} for fallback sensitivity.}
\end{table}

\paragraph{Content changes and missing submissions.}
The 320 static revision cells contain 68 final failures: 66 initially incorrect submissions remain incorrect, one correct answer becomes an incorrect answer, and one correct initial program is followed by an empty final submission.
The content-level loss is Qwen3-Next mathematics: both its initial and revised answers are nonempty and complete.
The submission loss is Flash programming: its critique exhausts its 131{,}072-token stage allowance before revision begins, and the harness discards the correct initial program.
Conversely, Flash's two mathematics W$\to$R transitions include one repair of a nonempty wrong answer and one recovery from an initial generation truncated at 131{,}072 tokens.
Thus the operational W$\to$R and R$\to$W counts do not alone identify content-level repair or damage.

\paragraph{Post-hoc sensitivity: retaining the initial answer.}
\label{app:revision-fallback}
We replay a deterministic fallback on all 320 static revision records: if the cell ends with a stage failure, truncation, or exhausted budget, has no final submission, and has a nonempty completed initial answer, return that initial answer.
The trigger uses only termination status and output availability; existing hidden grades score the returned answer but never determine whether fallback occurs.
Exactly one cell qualifies, the Flash programming case above, raising its accuracy from 98/100 to 99/100, equal to Direct.
All other static revision scores are unchanged, and all spent tokens and logical API costs remain charged.
This analysis uses saved outputs and grades without new inference or grading; the frozen primary scores remain unchanged.
It isolates one avoidable loss from the failure-handling rule and does not establish that revision preserves correctness in general.
The replay does not apply to agent episodes, where actions may already have changed the environment.

\paragraph{Candidate-pool replay.}
Every retained mathematics Parallel@4 record contains three graded candidates: 60 pools each for Flash, Qwen3-Next, and MiniMax, 53 for GLM-5, and 59 for GPT-OSS.
We apply the original final-answer parser and voting tie rule to these same strings, then look up the selected candidate's existing grade.
The choice rule has no access to correctness labels.
This isolates the choice rule on observed pools, while the eight missing records remain failures in the primary 60-task denominators.
On programming, 496 records contain all three attempt slots, including empty attempts; two GPT-OSS records are partial and two are missing. Coverage of partial pools is reported only when a correct candidate is observed.
No unavailable candidate is imputed correct or incorrect for the conditional conversion calculation.

\subsection{Per-model diagnostics behind Section~\ref{sec:results-rq2}}
\label{app:diagnostics}
\suppressfloats[t]

\paragraph{Mathematics.}
Table~\ref{tab:mathcoverage} grades every available three-candidate pool and applies an offline vote to exactly the same candidates used by the model judge.
Flash's pools contain a correct answer on 54/60 tasks, and self-selection converts 52 of those opportunities.
Qwen3-Next's pools cover 38/60 tasks, exceeding its Direct count of 31, yet its judge submits only 30 correct answers: eight covered pools are lost through an incorrect choice, with no missing choice.
On the 292 available pools across the five models, self-selection submits 185 correct answers and the counterfactual vote@3 submits 182, with ten wins and seven losses between them.
Flash contributes two additional successes under self-selection (52 versus 50), and Qwen3-Next submits one fewer (30 versus 31).
This replay holds the candidate pool fixed and changes only the choice rule, but it does not establish a broadly superior selector: the outcomes vary by model, and the model judge incurs additional inference cost.
Missing pools remain failures in the primary accuracy table and are excluded only from this explicitly conditional replay.

\begin{table}[t]
\centering\small
\caption{Mathematics Parallel@4: coverage, self-selection, and counterfactual vote@3 on the identical recorded pools. Every available pool has three graded candidates. Counts in the middle columns use the available-pool denominator; official accuracy always uses all 60 scheduled tasks.}
\label{tab:mathcoverage}
\begin{tabular}{@{}lrrrrr@{}}
\toprule
Model & Pools & Coverage & Self-select & Vote@3 & Conversion \\
\midrule
DeepSeek V4.1 Flash & 60 & 54 & 52 & 50 & 52/54 \\
Qwen3-Next-80B & 60 & 38 & 30 & 31 & 30/38 \\
GLM-5 & 53 & 44 & 41 & 41 & 41/44 \\
GPT-OSS-120B & 59 & 42 & 35 & 34 & 35/42 \\
MiniMax M2.5 & 60 & 31 & 27 & 26 & 27/31 \\
\bottomrule
\end{tabular}
\smallskip
\parbox{\linewidth}{\footnotesize GLM-5 and GPT-OSS have seven and one missing records, respectively. Voting is recomputed from stored final-answer strings with the existing parser and tie rule; it uses no hidden correctness labels to choose a candidate and is not an independently rerun protocol.}
\end{table}

\paragraph{Programming.}
Table~\ref{tab:stagefail} gives the corresponding decomposition for programming.
GLM-5 has a correct program in 88 pools but submits only 55; 27 covered pools return no usable selection and six select an incorrect program.
Flash loses three of its 99 covered pools, all without a usable choice.
Qwen3-Next exhibits a different failure profile: seven wrong choices and only one absent choice among its eight lost opportunities.
Across the five models, 81 observed covered pools fail to convert: 64 have no valid selection index and 17 select an incorrect candidate, including one empty answer. MiniMax accounts for that empty selected answer: its eight absent indices plus this case explain the nine submissions recovered by nonempty fallback (81 to 90).
Of the 61 coding outcomes lost against Direct, 56 have a retained record with no valid selection index, two have no final record, and three select an incorrect nonempty candidate.
This does not mean that every failure occurs inside the judge: GPT-OSS also has two partial candidate pools, and a missing record cannot be assigned a selection-stage cause.
It does show that the main loss in submitted accuracy cannot be described solely as choosing the wrong program after successful execution of the protocol.
The contrast with Parallel@2 is also not a clean selector ablation, because that protocol uses two candidates rather than three; the candidate-level accounting establishes the conversion loss without requiring that causal interpretation.

\begin{table}[t]
\centering\small
\caption{Programming Parallel@4: where recorded candidate coverage is lost. Each model has 100 scheduled tasks. $O$: a correct candidate is observed; $S$: the final submission is correct. The last two columns partition $O-S$. A missing record is not counted as an observed empty pool.}
\label{tab:stagefail}
\begin{tabular}{@{}lrrrrr@{}}
\toprule
Model & 3 attempts & $O$ & $S$ & No choice & Wrong choice \\
\midrule
DeepSeek V4.1 Flash & 100 & 99 & 96 & 3 & 0 \\
Qwen3-Next-80B & 100 & 83 & 75 & 1 & 7 \\
GLM-5 & 100 & 88 & 55 & 27 & 6 \\
GPT-OSS-120B & 96 & 94 & 69 & 25 & 0 \\
MiniMax M2.5 & 100 & 93 & 81 & 8 & 4 \\
\bottomrule
\end{tabular}
\smallskip
\parbox{\linewidth}{\footnotesize GPT-OSS has 96 pools with all three attempts recorded, two partial pools (both covered), and two missing records. ``No choice'' means no valid selection index; the selection stage need not have been reached. ``Wrong choice'' means an indexed candidate grades incorrect, including an empty answer. MiniMax has eight absent indices and one indexed empty answer (within its four wrong choices), so nonempty fallback recovers nine submissions. All other models record three attempts on 100 tasks; an attempt may be empty.}
\end{table}

The corresponding failure partition is shown in Figure~\ref{fig:conversion}.

\paragraph{Agents.}
On workflows, Flash's small gains under plan-level parallelism (78/160 to 82/160 and 81/160) coexist with losses for GLM-5, GPT-OSS, and MiniMax and near-floor performance for Qwen3-Next.
For workflow Parallel@2 and Parallel@4, GPT-OSS records 117/160 and 111/160 stage failures, respectively, and MiniMax records 159/160 under either protocol.
The recorded aggregate diagnostics associate these failures with absent final planning messages or tool-call markup in stages that require a plan or a selection.
These rows measure end-to-end performance of the model--protocol--interface combination, not the quality of a successfully executed multi-plan strategy.
Flash's workflow review lowers accuracy from 78/160 to 55/160 with 11 paired wins and 34 losses (Holm-adjusted $p=0.033$); because these are separate episodes, the 34 losses are not observations of an individual trajectory being repaired into failure.
On terminal tasks, Flash remains at 3/30 under either Parallel protocol, while Qwen3-Next and GLM-5 remain at zero; Flash nevertheless exchanges solved tasks with Direct, so unchanged accuracy does not mean that the protocols solve identical instances.
Thirteen recorded Flash terminal Direct cells and eight Opus terminal Direct cells terminate at a token or call limit, including cases whose final state is nevertheless correct.
Across the entire matrix, six resource-limited agent cells pass the native evaluator (Appendix~\ref{app:accounting}); retaining these grades distinguishes task completion from the harness's stopping condition.

\subsection{Same-pool selection and failure-handling replay}
\label{app:selector-replay}
\suppressfloats[t]

We compare choice rules on stored candidate strings, using hidden grades only after selection. This post-hoc analysis preserves the frozen primary matrix; it does not add generation samples, replace missing records, or estimate repeated-generation variability. We retain per-pool choices, public execution evidence, stage costs, and input hashes in the replay records; \texttt{analysis/run\_selector\_replay.py} and \texttt{analysis/summarize\_selector\_replay.py} reproduce execution and aggregation. Figure~\ref{fig:selector-replay-cost} in Section~\ref{sec:results-rq2} plots the resulting accuracy against logical API cost.

\paragraph{Programming rules and execution.}
We process all 498 retained pools: 496 pools with all three attempt slots recorded and two partial GPT-OSS pools. Two missing GPT-OSS records remain wrong in the denominator of 500 scheduled cells. We compare the frozen judge (J), fallback to the first nonempty candidate when J returns no valid nonempty choice (J+F), public-example fallback under the same condition (J+P), and public-example selection on all three candidates (P3) or only the first two of the same pool (P2). Public selection maximizes examples passed and breaks ties by candidate order; an empty or uncompilable candidate receives a negative score. No hidden grade determines selection or fallback. These pools retain empty or truncated attempts as failures; they do not require three usable programs. As further availability-conditioned checks, the 459 pools with three nonempty answers give 354 correct submissions for J versus 423 for P3. The 419 pools where all three candidates compile give 336 versus 392. These descriptive subsets do not replace the scheduled denominator.

Public execution reuses the frozen grading helper and immutable runtime image with network disabled, read-only containers, bounded memory, and the original compilation and execution limits. The replay makes 3{,}959 public-test executions in 838 seconds with four workers (3{,}299 summed worker-seconds), with no environment errors. Every candidate recorded as hidden-correct passes its public examples. No hidden tests are re-executed, and no new API request is made. These runtime measurements report the local execution expense separately from API dollars.

\paragraph{A public-first judge gate.}
An additional post-hoc rule (G) returns a unique public-score winner directly. For tied nonnegative maxima, it reuses the original judge's choice if that choice is among the maxima, otherwise returns the earliest maximum. If no candidate has a nonnegative score it returns no submission. The rule was recorded before inspecting aggregate public-test results. It neither uses hidden labels to decide whether to judge nor changes the judge prompt: the retained judge saw the original full pool, not newly supplied public scores or a filtered candidate list. This is a conditional replay of that recorded call, not an evaluation of a newly prompted judge. G calls the judge on 465 of 498 retained pools; its choices equal those of always judging with the same public-score guard, but its logical cost omits judging on unique winners.

\begin{table}[t]
\centering\small
\caption{Programming selector replay on the recorded Parallel@4 pools. Counts use 100 scheduled tasks per model. J: frozen judge; J+F: first-nonempty fallback; J+P: public-example fallback (both also handle an empty chosen answer); P3/P2: public-example choice on all three/first two candidates; G: public tests first, judge only tied maxima, with public fallback.}
\label{tab:selector-replay-code}
\setlength{\tabcolsep}{4pt}
\begin{tabular*}{\linewidth}{@{\extracolsep{\fill}}lrrrrrrr@{}}
\toprule
Model & Direct & J & J+F & J+P & P3 & G & P2 \\
\midrule
Flash & 99 & 96 & 99 & 99 & 99 & 99 & 99 \\
Qwen3-Next & 76 & 75 & 76 & 76 & 82 & 81 & 82 \\
GLM-5 & 66 & 55 & 78 & 81 & 87 & 87 & 83 \\
GPT-OSS & 91 & 69 & 94 & 94 & 94 & 94 & 94 \\
MiniMax & 87 & 81 & 90 & 90 & 91 & 91 & 90 \\
Total & 419 & 376 & 437 & 440 & 453 & 452 & 448 \\
Choice indices & --- & 421 & 498 & 497 & 497 & 497 & 493 \\
\bottomrule
\end{tabular*}
\smallskip
\parbox{\linewidth}{\footnotesize All rules use public information only. Two missing GPT-OSS records remain failures. Its two partial pools are retained where a candidate exists; 496 pools with three recorded attempts are analyzed separately; some attempts are empty. Choice indices count a selected index, not guaranteed correct or nonempty output (one J choice is empty). The index-only fallbacks score 436 and 439; output-aware J+F/J+P also recover one empty choice. Both retain all judge cost; G calls the judge only for tied maxima and guards its choice by public scores; P3/P2 omit judge calls. P2 reuses the first two candidates of these pools, not the independently recorded Parallel@2 runs.}
\end{table}

\begin{table}[t]
\centering\small
\caption{Logical API costs of the coding replay (USD per retained task). Fallback retains all judge spending; public-only variants omit it, and G charges the judge only for public-score ties.}
\label{tab:selector-replay-code-cost}
\setlength{\tabcolsep}{4pt}
\begin{tabular*}{\linewidth}{@{\extracolsep{\fill}}lrrrr@{}}
\toprule
Model & J / J+F / J+P & P3 & P2 & G \\
\midrule
Flash & 0.0766 & 0.0391 & 0.0269 & 0.0766 \\
Qwen3-Next & 0.0800 & 0.0566 & 0.0381 & 0.0792 \\
GLM-5 & 0.3774 & 0.3058 & 0.2096 & 0.3647 \\
GPT-OSS & 0.0809 & 0.0709 & 0.0533 & 0.0793 \\
MiniMax & 0.1686 & 0.1172 & 0.0750 & 0.1646 \\
\bottomrule
\end{tabular*}
\smallskip
\parbox{\linewidth}{\footnotesize Candidate generation includes its original public-tool interactions. Public-test replay itself makes no model calls; its local execution time is reported separately and is not priced as zero-cost computation. These are accounting counterfactuals on saved pools, not fresh end-to-end budget trials.}
\end{table}

\paragraph{Accuracy and selection cost.}
The original index-only fallback recovers 60 correct submissions (436/500). Inspection revealed one further case in which a valid judge index refers to an empty candidate. The output-aware amendment triggers fallback for this public condition as well, giving J+F 437/500 and J+P 440/500, compared with 439/500 for index-only public fallback. This additional rule was specified after that inspection and is reported as post-hoc. J+F therefore recovers 61 submissions at unchanged logical cost, changing the gain against Direct from $-8.6$ to $+3.6$ pp. P3 yields 453/500 ($+6.8$ pp against Direct), converts 453 of 457 covered pools, and saves 12--49\% of logical API cost across models. P2 yields 448/500, so five extra correct submissions are attributable to adding the third candidate under this fixed-pool rule. The independently generated Parallel@2 score of 450/500 is a different comparison. G yields 452/500: in these observations the extra judging adds cost without an accuracy gain over P3.

\paragraph{Selection among mixed candidates.}
\label{app:mixed-pools}
Table~\ref{tab:selector-composition} separates pools by the correctness of their recorded candidates. Of 498 retained pools, 359 contain only correct candidates, 41 only incorrect candidates, and 98 contain both. The first group includes the two partial pools; it does not require three recorded attempts. In the mixed group, the judge, nonempty fallback, and public-example selection obtain 61, 78, and 94 correct submissions, respectively. Public selection gains 16 choices and loses none against fallback, accounting for its entire additional gain. Thus it improves selection within mixed pools as well as preserving available answers. These groups are defined retrospectively for diagnosis; hidden grades never determine the selection rule.

\begin{table}[t]
\centering\small
\caption{Programming conversion by recorded candidate correctness. J: frozen judge; J+F: nonempty fallback; P3: public-example selection. Counts exclude the two missing records, which remain failures in scheduled accuracy.}
\label{tab:selector-composition}
\begin{tabular}{@{}lrrrr@{}}
\toprule
Candidate pool & Pools & J & J+F & P3 \\
\midrule
All correct & 359 & 315 & 359 & 359 \\
Mixed & 98 & 61 & 78 & 94 \\
All incorrect & 41 & 0 & 0 & 0 \\
\bottomrule
\end{tabular}
\smallskip
\parbox{\linewidth}{\footnotesize The two partial pools are included in the all-correct group. These are descriptive groups, not information supplied to selectors. P3 versus J+F on mixed pools has 16 wins and no losses.}
\end{table}

Among the 465 pools with tied highest public scores, the tied candidates are all correct in 426 pools, all incorrect in 26, and mixed in 13. Considering every candidate in those same pools instead gives 358 all-correct, 26 all-incorrect, and 81 mixed pools. The distinction matters: lower-scoring incorrect candidates need not remain in the tied set. Public selection submits correctly on 435 of these 465 pools, including nine of the 13 with mixed tied maxima. This explains why frequent public-score ties need not imply frequent consequential ambiguity, and why the tested judge gate adds little value.

All candidate-generation calls, including their original tool interactions, remain charged. Fallback after judging retains every spent judge-stage call; policies that omit judging remove all calls in that stage, including continuations. G charges the recorded judge stage only on ties. Per-call sums reconcile with each original cell's logical cost. These are accounting counterfactuals on fixed pools, not refunds of historical spending or fresh end-to-end trials at matched expenditure.

\paragraph{Mathematics.}
Vote@3 uses the original parser and earliest-candidate tie rule on all 292 retained complete pools. Original judging obtains 185 correct submissions versus 182 for voting, with ten paired wins and seven losses. Applying the same nonempty-answer fallback to all 292 mathematics pools recovers one MiniMax answer after an invalid judge choice, raising MiniMax from 27 to 28 and the total to 186, at unchanged cost. All other models remain unchanged. Removing all judge-stage calls saves 12.3--21.1\% of logical API cost across models (Table~\ref{tab:selector-replay-math-cost}); generation remains charged. Flash gains two correct answers with judging at an additional \$0.032 per task, whereas Qwen3-Next voting is cheaper and one answer more accurate.

\begin{table}[t]
\centering\small
\caption{Mathematics original judging (J), judging with nonempty fallback (J+F), and vote@3 on identical pools. Costs are mean logical API dollars per retained task at the original price basis; voting removes the entire judge stage.}
\label{tab:selector-replay-math-cost}
\setlength{\tabcolsep}{4pt}
\begin{tabular*}{\linewidth}{@{\extracolsep{\fill}}lrrrrrrr@{}}
\toprule
Model & Pools & J & J+F & Vote & J / J+F \$ & Vote \$ & Saving \\
\midrule
Flash & 60 & 52 & 52 & 50 & 0.1764 & 0.1444 & 18.1\% \\
Qwen3-Next & 60 & 30 & 30 & 31 & 0.1561 & 0.1369 & 12.3\% \\
GLM-5 & 53 & 41 & 41 & 41 & 0.6183 & 0.4882 & 21.1\% \\
GPT-OSS & 59 & 35 & 35 & 34 & 0.0959 & 0.0758 & 21.0\% \\
MiniMax & 60 & 27 & 28 & 26 & 0.1511 & 0.1193 & 21.0\% \\
\bottomrule
\end{tabular*}
\smallskip
\parbox{\linewidth}{\footnotesize Missing pools are excluded from this controlled comparison, while remaining failures in the frozen matrix. Fallback recovers one MiniMax answer (27 to 28), giving 186 correct versus 185 for J and 182 for voting; all judge cost remains charged. Choice rules use no hidden correctness labels.}
\end{table}

\paragraph{Mathematics grading sensitivity.}
\label{app:grading-sensitivity}
We keep the candidates and selected indices fixed and score all 292 retained pools using their saved CompassVerifier-3B candidate judgments (Table~\ref{tab:selector-grading}). The verifier's source hashes and candidate order are checked against the original evaluations; no new generation or grading is performed. The primary rule grader gives 186 correct submissions for judging with fallback and 182 for voting. Compass gives 194 and 189, respectively: 11 paired wins and six losses, or $+1.71$ pp with a pointwise 95\% interval of $[-1.04,4.79]$. Two of the 16 pairs discordant under the primary grader have a selected-answer grading disagreement. The observed direction is preserved, while both graders leave uncertainty about the small selector difference. We retain the frozen primary scores; this is an auxiliary sensitivity analysis, not human adjudication or selection of the more favorable grader. The task-stratified bootstrap uses 20{,}000 replicates and seed 20260926.

\begin{table}[t]
\centering\small
\caption{Mathematics selector sensitivity under two grading rules on identical candidates and fixed choices. J: frozen judge; J+F: nonempty fallback; V: vote@3. W--L compares J+F with V.}
\label{tab:selector-grading}
\begin{tabular}{@{}lrrrrr@{}}
\toprule
Grader & Pools & J & J+F & V & W--L \\
\midrule
Primary rule & 292 & 185 & 186 & 182 & 10--6 \\
Compass & 292 & 192 & 194 & 189 & 11--6 \\
\bottomrule
\end{tabular}
\smallskip
\parbox{\linewidth}{\footnotesize Compass judgments were saved during the evaluation campaign. This auxiliary comparison does not replace the primary grader or constitute human adjudication.}
\end{table}

The script \texttt{analysis/audit\_selector\_sensitivity.py} reproduces the candidate-composition and grading-sensitivity analyses from saved records, including per-model counts and source hashes.

\paragraph{Deployment comparison after replay.}
We add vote@3, mathematics nonempty fallback, and all coding replay variants to the original configurations and recompute nondominance using scheduled-task accuracy and mean logical API cost over retained records (\texttt{analysis/audit\_replay\_deployment.py}). The peak-price mathematics and coding frontiers are unchanged. At half the Flash price, vote@3 adds a frontier point at 50/60 and \$0.0722/task, below judging's 52/60 and \$0.0882; Opus remains at 57/60 and \$0.1035. On code, Flash Direct remains the frontier at either price basis. These point-estimate, API-cost comparisons omit local execution expense and do not establish statistically certified dominance.

\paragraph{Paired uncertainty and scope.}
We bootstrap tasks within the frozen strata, keeping the five models together (20{,}000 replicates, seed 20260926). These post-hoc pointwise intervals describe the fixed panel and saved pools; they are not simultaneous significance tests. On all 500 scheduled coding cells, P3 versus J gains $15.4$ pp with interval $[12.6,18.2]$ (77 wins, no losses). Restricting both to the 496 pools with three recorded attempts gives $15.3$ pp, $[12.7,18.1]$ (76 wins, no losses). P3 versus same-pool P2 gains $1.0$ pp, $[0.2,2.0]$ (five wins, no losses); G versus P3 changes $-0.2$ pp, $[-0.6,0.0]$ (no wins, one loss). Conditional on the 292 retained mathematics pools, original judging versus voting gains $1.0$ pp, $[-1.7,3.8]$; judging with nonempty fallback gains $1.4$ pp, $[-1.4,4.1]$ (10 wins, 6 losses). None of these comparisons estimates variability from new generations or repeats the agent environments.

\clearpage  
\section{Full results}
\label{app:full}

Table~\ref{tab:main} complements Table~\ref{tab:summary} in Section~\ref{sec:results-rq1}: for every configuration it adds the paired wins--losses against the same model's Direct run and the cost relative to Direct.

\begin{table*}[!ht]
\centering
\small
\setlength{\tabcolsep}{3.5pt}
\caption{Frozen protocol matrix: accuracy of each cheap model under each protocol, with wins--losses against the same model's Direct on the same tasks (exact two-sided sign test: $^{*}p<0.05$, $^{**}p<0.01$, uncorrected) and cost relative to Direct. The rightmost column gives the expensive anchor, Claude Opus 5.5, answering once. n/a: excluded by the configured revision headroom requirement. --: not run. $^\ddagger$: includes cells that never finished, scored incorrect. $^\S$: the plan or plan-selection step failed its required format in at least 69\% of tasks, so the row reflects format compliance rather than the value of extra compute.}
\label{tab:main}
\resizebox{\linewidth}{!}{%
\begin{tabular}{@{}ll rr rr rr rr r@{}}
\toprule
 & & \multicolumn{2}{c}{Direct} & \multicolumn{2}{c}{Parallel@2} & \multicolumn{2}{c}{Parallel@4} & \multicolumn{2}{c}{Revise@4} & Anchor \\
\cmidrule(lr){3-4} \cmidrule(lr){5-6} \cmidrule(lr){7-8} \cmidrule(lr){9-10} \cmidrule(l){11-11}
Domain & Model & Acc & \$/task & Acc (W--L) & Cost & Acc (W--L) & Cost & Acc (W--L) & Cost & Opus 5.5 \\
\midrule
\multirow{5}{*}{Math (60)} & DeepSeek V4.1 Flash & 0.767 & 0.051 & 0.783 (1--0) & 1.9$\times$ & 0.867 (6--0)$^{*}$ & 3.4$\times$ & 0.800 (2--0) & 1.9$\times$ & \multirow{5}{*}{0.950} \\
 & GPT-OSS-120B & 0.533 & 0.026 & 0.533 (0--0) & 2.0$\times$ & 0.583$^\ddagger$ (5--2) & 3.7$\times$ & n/a & n/a &  \\
 & MiniMax M2.5 & 0.417 & 0.038 & 0.417 (0--0) & 2.0$\times$ & 0.450 (3--1) & 3.9$\times$ & n/a & n/a &  \\
 & GLM-5 & 0.650 & 0.175 & 0.667$^\ddagger$ (2--1) & 2.0$\times$ & 0.683$^\ddagger$ (7--5) & 3.5$\times$ & n/a & n/a &  \\
 & Qwen3-Next-80B & 0.517 & 0.048 & 0.517 (0--0) & 1.9$\times$ & 0.500 (3--4) & 3.3$\times$ & 0.500 (0--1) & 1.2$\times$ &  \\
\midrule
\multirow{5}{*}{Code (100)} & DeepSeek V4.1 Flash & 0.990 & 0.013 & 0.990 (0--0) & 2.0$\times$ & 0.960 (0--3) & 5.8$\times$ & 0.980 (0--1) & 1.3$\times$ & \multirow{5}{*}{0.930\,(0.990$^\dagger$)} \\
 & GPT-OSS-120B & 0.910 & 0.025 & 0.970 (6--0)$^{*}$ & 2.0$\times$ & 0.690$^\ddagger$ (4--26)$^{**}$ & 3.2$\times$ & n/a & n/a &  \\
 & MiniMax M2.5 & 0.870 & 0.039 & 0.910 (4--0) & 1.9$\times$ & 0.810 (2--8) & 4.3$\times$ & n/a & n/a &  \\
 & GLM-5 & 0.660 & 0.103 & 0.810 (15--0)$^{**}$ & 2.1$\times$ & 0.550 (10--21) & 3.7$\times$ & n/a & n/a &  \\
 & Qwen3-Next-80B & 0.760 & 0.019 & 0.820 (6--0)$^{*}$ & 2.0$\times$ & 0.750 (2--3) & 4.2$\times$ & 0.760 (0--0) & 1.3$\times$ &  \\
\midrule
\multirow{5}{*}{Workflow (160)} & DeepSeek V4.1 Flash & 0.487 & 0.020 & 0.512 (34--30) & 1.0$\times$ & 0.506 (35--32) & 1.3$\times$ & 0.344 (11--34)$^{**}$ & 1.5$\times$ & \multirow{5}{*}{0.925} \\
 & GPT-OSS-120B & 0.094 & 0.015 & 0.019$^\S$ (2--14)$^{**}$ & 0.2$\times$ & 0.019$^\S$ (3--15)$^{**}$ & 0.3$\times$ & n/a & n/a &  \\
 & MiniMax M2.5 & 0.119 & 0.094 & 0.006$^\S$ (1--19)$^{**}$ & $<$0.1$\times$ & 0.000$^\S$ (0--19)$^{**}$ & $<$0.1$\times$ & n/a & n/a &  \\
 & GLM-5 & 0.138 & 0.129 & 0.069 (9--20) & 0.6$\times$ & 0.062 (7--19)$^{*}$ & 0.4$\times$ & n/a & n/a &  \\
 & Qwen3-Next-80B & 0.000 & 0.097 & 0.019 (3--0) & 0.8$\times$ & 0.019 (3--0) & 0.8$\times$ & 0.000 (0--0) & 0.5$\times$ &  \\
\midrule
\multirow{5}{*}{Terminal (30)} & DeepSeek V4.1 Flash & 0.100$^\ddagger$ & 0.307 & 0.100$^\ddagger$ (3--3) & 1.2$\times$ & 0.100 (2--2) & 1.4$\times$ & 0.000 (0--3) & 0.7$\times$ & \multirow{5}{*}{0.433} \\
 & GPT-OSS-120B & 0.000 & 0.175 & -- & -- & -- & -- & n/a & n/a &  \\
 & MiniMax M2.5 & 0.000 & 0.441 & -- & -- & -- & -- & n/a & n/a &  \\
 & GLM-5 & 0.000 & 0.350 & 0.000 (0--0) & 0.6$\times$ & 0.000 (0--0) & 0.6$\times$ & n/a & n/a &  \\
 & Qwen3-Next-80B & 0.000 & 0.063 & 0.000 (0--0) & 0.9$\times$ & 0.000 (0--0) & 0.7$\times$ & 0.000 (0--0) & 0.6$\times$ &  \\
\bottomrule
\end{tabular}}
\\[2pt]{\footnotesize $^\dagger$Opus 5.5 scores 0.990 on code when numpy is available in the grading sandbox; six of its programs import it. Costs average retained records at configured rates (DeepSeek peak); missing records have no imputed cost. Holm-adjusted tests are in Table~\ref{tab:holm}.}
\end{table*}

\end{document}